%% file: FastUnmixing.tex
\documentclass[10pt,journal,final,a4paper,twoside,twocolumn]{IEEEtran}
\usepackage{dsfont}
\usepackage{amsmath}
\usepackage{amssymb}
\usepackage{amsfonts}
\usepackage{graphicx}
\usepackage{amsmath}
\usepackage[all,poly]{xy}
\usepackage{multirow}
\usepackage{color}
\usepackage[noadjust]{cite}
\usepackage{subfigure}
\usepackage{mathrsfs}
\usepackage{algorithmic}
\usepackage[linesnumbered,lined,ruled]{algorithm2e}
\usepackage{cancel}
\usepackage{url}
\usepackage{lipsum}
\usepackage{stfloats}
\usepackage{epsfig}
\usepackage{bm,balance}
\usepackage{amsthm}
\usepackage{gensymb}
\usepackage{url}

\usepackage{epstopdf}

\usepackage{setspace}

\newtheorem{theorem}{Theorem}

\newtheorem*{remark*}{Remark}

\newcommand{\bs}{\boldsymbol}

\newcommand{\tcpp}[1]{\tcp{\emph{\small{#1}}}}

\input notations.tex

\graphicspath{{figures/}}

\title{Fast Spectral Unmixing based on Dykstra's Alternating Projection}

\author{\IEEEauthorblockN{Qi Wei, \IEEEmembership{Student Member,~IEEE},
Jos\'e Bioucas-Dias, \IEEEmembership{Member,~IEEE},\\
Nicolas Dobigeon, \IEEEmembership{Senior Member,~IEEE}, and Jean-Yves
Tourneret, \IEEEmembership{Senior Member,~IEEE}}
\thanks{Part of this work has been supported by
the Chinese Scholarship Council, the Hypanema ANR Project n$^\circ$ANR-12-BS03-003,
the ANR-11-LABX-0040-CIMI Project, in particular during the ANR-11-IDEX-0002-02 program within
the thematic trimester on image processing, and the Portuguese Science and Technology Foundation
under Projects {UID/EEA/50008/2013} and PTDC/EEI-PRO/1470/2012.}
\thanks{Qi Wei, Nicolas Dobigeon and Jean-Yves Tourneret are with University of Toulouse, IRIT/INP-ENSEEIHT, 31071 Toulouse cedex 7, France (e-mail: \{qi.wei, nicolas.dobigeon, jean-yves.tourneret\}@enseeiht.fr).}
\thanks{Jos\'e Bioucas-Dias is with Instituto de Telecomunica\c{c}\~{o}es and Instituto
Superior T\'ecnico, Universidade de Lisboa, Portugal (e-mail: bioucas@lx.it.pt).}
}

\begin{document}

\maketitle

\begin{abstract}
This paper presents a fast spectral unmixing algorithm based on Dykstra's alternating projection. The proposed algorithm formulates the fully constrained least squares optimization problem associated with the spectral unmixing task as an unconstrained regression problem followed by a projection onto the intersection of several closed convex sets. This projection is achieved by iteratively projecting onto each of the convex sets individually, following Dyktra's scheme. The sequence thus obtained is guaranteed to converge to the sought projection.
Thanks to the preliminary matrix decomposition and variable substitution, the projection is implemented intrinsically in a subspace, whose dimension is very often much lower than the number of bands. A benefit of this strategy is that the order of the computational complexity for each projection is decreased from quadratic to linear time. Numerical experiments considering diverse spectral unmixing scenarios provide evidence that the proposed algorithm competes with the state-of-the-art, namely  when the number of endmembers is relatively small, a circumstance often observed in real hyperspectral applications.

\end{abstract}


\section{Introduction}
\label{sec:intro}
\input{introduction.tex}


\section{Proposed Fast Unmixing Algorithm}
\label{sec:proposed}
\input{proposed.tex}

\section{Experiments using Synthetic and Real Data}
\label{sec:simu}
\input{simulation.tex}

\section{Conclusion}
\label{sec:concls}
This paper proposed a fast unmixing method based on an alternating projection strategy.
Formulating the spectral unmixing problem as a projection onto the intersection of convex
sets allowed Dykstra's algorithm to be used to compute the solution of this unmixing problem.
The projection was implemented intrinsically in a subspace, making the proposed algorithm
computationally efficient. In particular, the proposed unmixing algorithm showed similar 
performance comparing to state-of-the-art methods, with significantly reduced execution time,
especially when the number of endmembers is small or moderate, which is often the case
when analyzing conventional multi-band images.
Future work includes the generalization of the proposed algorithm to cases
where the endmember matrix is rank deficient or ill-conditioned.

\vfil

\appendix[Solving \eqref{eq:opt_ASC_ANC_a} with KKT conditions]
\label{App:A}
\input{appendix_A.tex}

\section*{Acknowledgments}
The authors would like to thank Rob Heylen, \'{E}milie Chouzenoux and Sa\"{i}d Moussaoui
for sharing the codes of \cite{Heylen2013,Chouzenoux2014} used in our experiments. They are
also grateful to Nathalie Brun for sharing the EELS data and offering useful suggestions to process them.
They would also thank Frank Deutsch for helpful discussion on the convergence rate of 
the alternating projections.

\bibliographystyle{ieeetran}
\bibliography{strings_all_ref,biblio_all}
\end{document}

%% file: notations.tex
\input com_notations.tex

\newcommand{\apriori}{\emph{a priori }}

\newcommand{\argmin}{\mathrm{arg}\min}

\newcommand{\Id}[1]{\textbf{I}_{#1}}

\newcounter{algo}
\renewcommand{\thealgo}{\arabic{algo}}

%% file: com_notations.tex
\def\bfb{{\mathbf{b}}}
\def\bfc{{\mathbf{c}}}
\def\bfd{{\mathbf{d}}}
\def\bfe{{\mathbf{e}}}

\def\bfs{{\mathbf{s}}}

\def\bfu{{\mathbf{u}}}
\def\bfv{{\mathbf{v}}}
\def\bfw{{\mathbf{w}}}

\def\bfy{{\mathbf{y}}}
\def\bfz{{\mathbf{z}}}

\def\bfA{{\mathbf{A}}}

\def\bfD{{\mathbf{D}}}
\def\bfE{{\mathbf{E}}}

\def\bfI{{\mathbf{I}}}

\def\bfN{{\mathbf{N}}}

\def\bfP{{\mathbf{P}}}
\def\bfQ{{\mathbf{Q}}}

\def\bfU{{\mathbf{U}}}
\def\bfV{{\mathbf{V}}}

\def\bfX{{\mathbf{X}}}
\def\bfY{{\mathbf{Y}}}
\def\bfZ{{\mathbf{Z}}}

\def\calS{{\mathcal{S}}}

\def\calA{{\mathcal{A}}}

\def\calC{{\mathcal{C}}}

\def\calJ{{\mathcal{J}}}

\def\calN{{\mathcal{N}}}

\def\calS{{\mathcal{S}}}

\def\bsa{{\boldsymbol{a}}}

\def\bsx{{\boldsymbol{x}}}

%% file: introduction.tex
\IEEEPARstart{S}{pectral} unmixing (SU) aims at decomposing a set of $n$ multivariate measurements $\bfX = \left[\bsx_1,\ldots,\bsx_{n}\right]$ into a collection of $m$ elementary signatures $\bfE=\left[\bfe_1,\cdots,\bfe_m\right]$, usually referred
to as \emph{endmembers}, and estimating the relative proportions $\bfA=\left[\bsa_1,\ldots,\bsa_n\right]$ of these
signatures, called \emph{abundances}. SU has been advocated as a relevant multivariate analysis technique in various applicative areas, including remote sensing \cite{Averbuch2012}, planetology \cite{Themelis2012pss}, microscopy \cite{Dobigeon2012ultra}, spectroscopy \cite{Carteret2009} and gene expression analysis \cite{Huang2011plos}. In particular, it has demonstrated a great interest when analyzing multi-band (e.g., hyperspectral) images, for instance for pixel
classification \cite{Chang1998b}, material quantification \cite{Wang2006} and subpixel detection
\cite{Manolakis2001}.

In this context, several models have been proposed in the literature to properly describe the physical process underlying the observed measurements. Under some generally mild assumptions \cite{Dobigeon2014}, these measurements are supposed to result from linear combinations of the elementary spectra, according to the popular \emph{linear mixing model} (LMM) \cite{Keshava2002,Bioucas-Dias2012, Ma2014signal}. More precisely,
each column $\bsx_j \in \mathbb{R}^{n_{\lambda}}$ of the
measurement matrix $\bfX = \left[\bsx_1,\ldots,\bsx_{n}\right]$
can be regarded as a noisy linear combination of the spectral signatures leading to the following matrix formulation
\begin{equation}
\begin{array}{ll}
\label{eq:obs_general}
\bf X =  EA + N\\
\end{array}
\end{equation}
where
\begin{itemize}
\item $\bfE \in\mathbb{R}^{n_{\lambda} \times m}$ is the endmember matrix whose columns $\bfe_1,\cdots,\bfe_m$
are the signatures of the $m$ materials,
\item $\bfA \in \mathbb{R}^{m \times n}$ is the abundance matrix whose $j$th column $\bsa_j \in \mathbb{R}^{m}$
contains the fractional abundances of the $j$th spectral vector $\bsx_j$, 
\item $\bfN \in \mathbb{R}^{n_{\lambda} \times n}$ is the additive noise matrix.
\end{itemize}

As the mixing coefficient $a_{i,j}$ represents the proportion (or probability of occurrence) of the the $i$th endmember in
the $j$th measurement \cite{Keshava2002,Bioucas-Dias2012}, the abundance vectors satisfy the following \emph{abundance non-negativity constraint} (ANC)
and \emph{abundance sum-to-one constraint} (ASC)
\begin{equation}
\label{eq:asc_anc_vector}
\bsa_j \geq 0 \quad \textrm{and} \quad \bs{1}_m^T \bsa_j=1, \forall j = 1,\cdots,n
\end{equation}
where $\geq$ means element-wise greater  {or equal} and $\bs{1}_m^T \in \mathbb{R}^{m \times 1}$ represents
a vector with all ones. Accounting for all the image pixels, the constraints \eqref{eq:asc_anc_vector}
can be rewritten in matrix form
\begin{equation}
\bfA\geq 0 \quad \textrm{and} \quad \bs{1}_m^T \bfA=\bs{1}_n^T.
\end{equation}

Unsupervised linear SU boils down to estimating the endmember matrix $\bfE$ and abundance matrix $\bfA$ from the measurements $\bfX$ following the LMM \eqref{eq:obs_general}. It can be regarded as a special instance of
(constrained) blind source separation, where the endmembers are the sources \cite{Schmidt2010}. There already exists a lot of
algorithms for solving SU (the interested reader is invited to consult {\cite{Keshava2002,Bioucas-Dias2012,Ma2014signal}} for comprehensive reviews on the SU problem and existing unmixing methods). Most of the unmixing techniques tackle the SU problem into two successive steps. First, the endmember signatures are identified thanks to a prior knowledge regarding the scene of interest, or extracted from the data directly using dedicated algorithms, such as N-FINDR \cite{Winter1999spie}, vertex component analysis (VCA) \cite{Nascimento2005}, {and successive  volume maximization (SVMAX)
\cite{Chan2011simplex}. } Then, in a second step, called \emph{inversion} or \emph{supervised} SU, the
abundance matrix $\bfA$ is estimated given the previously identified endmember matrix $\bfE$, which is the problem addressed in this paper.

Numerous inversion algorithms have been developed in the literature, mainly based on deterministic or statistical approaches. Heinz \emph{et al.} \cite{Heinz2001} developed
a fully constrained least squares (FCLS) algorithm by generalizing the Lawson-Hanson
non-negativity constrained least squares (NCLS) algorithm \cite{Lawson1974}.
Dobigeon \emph{et al.} formulated
the unmixing problem into a Bayesian framework and proposed to draw samples from the
posterior distribution using a Markov chain Monte Carlo algorithm  \cite{Dobigeon2008}. This
simulation-based method considers the ANC and ASC both strictly while
the computational complexity is significant when compared with
other optimization-based methods. Bioucas-Dias \emph{et al.}
developed a sparse unmixing algorithm by variable splitting and augmented Lagrangian (SUnSAL)  and its constrained version (C-SUnSAL), which generalizes the
unmixing problem by introducing spectral sparsity explicitly \cite{Bioucas2010SUNSAL}.
More recently, Chouzenoux  \emph{et al.} \cite{Chouzenoux2014} proposed a primal-dual
interior-point optimization algorithm allowing for a constrained least squares (LS)
estimation approach and an algorithmic structure suitable for a parallel
implementation on modern intensive computing devices such as graphics processing
units (GPU). Heylen \emph{et al.} \cite{Heylen2013} proposed a new algorithm based on the
Dykstra's algorithm \cite{Boyle1986} for projections onto convex sets (POCS), with runtimes
that are competitive compared to several other techniques.

In this paper, we follow a Dykstra's strategy for POCS to solve the unmixing problem.
Using an appropriate decomposition of the endmember matrix and a variable substitution,
the unmixing problem is formulated as a projection onto the intersection of $m+1$
convex sets (determined by ASC and ANC) in a subspace, whose dimension is much
lower than the number of bands. The intersection of $m+1$ convex sets is
split into the intersection of $m$ convex set pairs, which guarantees that the
abundances always live in the hyperplane governed by ASC to accelerate the
convergence of iterative projections. In each projection, the subspace transformation
yields linear order (of the number of endmembers) computational operations which
decreases the complexity greatly when compared with Heylen's method \cite{Heylen2013}.

The paper is organized as follows. In Section \ref{sec:proposed}, we formulate
SU as a projection problem onto the intersection of convex sets defined
in a subspace with reduced dimensionality. We present the proposed
strategy for splitting the intersection of $m+1$ convex sets into the
intersection of $m$ convex set pairs.
Then, the Dykstra's alternating projection is used to solve this projection
problem, where each individual projection can be solved analytically.
The convergence and complexity analysis of the resulting algorithm
is also studied. Section \ref{sec:simu} applies the proposed algorithm to
synthetic and real multi-band data. Conclusions and future work are
summarized in Section \ref{sec:concls}.

%% file: proposed.tex
In this paper, we address the problem of supervised SU, which consists of solving the following optimization problem
\begin{align}
  \label{eq:fcld}
  \begin{aligned}
  &\min_{\bf A}  \Vert{{\bf X - E A}}\Vert_F^2\\
   & \text{subject to (s.t.)}  \quad \textbf{A}\geq 0 \quad \textrm{and} \quad{\bf 1}^T_m{\bf A} = {\bf 1}_n^T
   \end{aligned}
\end{align}
where $\|\cdot\|_F$ is the Frobenius norm. As explained in the introduction, this problem has been
considered in many applications where spectral unmixing plays a relevant role.
It is worthy to interpret this optimization problem from a probabilistic
point of view. The quadratic objective function can be easily related to the negative log-likelihood function associated with observations $\bfX$ corrupted by an additive white Gaussian noise. Moreover, the ANC and ASC constraints can be regarded as a
uniform distribution for $\bsa_j$ ($\forall j = 1,\cdots,n$) on the feasible region $\mathcal{A}$
\begin{equation}
p(\bsa_j)=\left\{
\begin{array}{ll}
c & \textrm{if } \bsa_j\in \mathcal{A} \\
0 & \textrm{elsewhere} \end{array} \right.
\end{equation}
where $\mathcal{A} = \left\{\bsa|\bsa \geq 0, \bs{1}^T \bsa=1 \right\}$ and $c=1\slash\mathrm{vol}(\mathcal{A})$. Thus, minimizing \eqref{eq:fcld} 
can be interpreted as maximizing the posterior distribution of $\bfA$ with the prior $p(\bfA)=\prod\limits_{j=1}^{n} p(\bsa_j)$, where we have 
assumed the abundance vectors $\bsa_i$ are \apriori independent.
In this section, we will demonstrate that the optimization problem \eqref{eq:fcld} can be decomposed into an
unconstrained optimization, more specifically an unconstrained least square (LS) problem with an explicit
closed form solution, followed by a projection step that can be efficiently achieved
with the Dykstra's alternating projection algorithm.

\subsection{Reformulating Unmixing as a Projection Problem}

Under the assumption that $\bfE$ has full column rank\footnote{This assumption is satisfied once the endmember spectral signatures are linearly independent.}, it is straightforward to show that the problem \eqref{eq:fcld} is
equivalent to
\begin{align}
\begin{aligned}
  &\min\limits_{\bf A}  \|{\bf Y - DA	}\|_F^2\\
  & \textrm{s.t.}  \quad\bfA \geq 0 \quad \textrm{and} \quad{\bf 1}^T_m{\bf A} = {\bf 1}_n^T
\end{aligned}
\label{eq:fcls_reduced_size}
\end{align}
where $\bfD$ is any ${m\times m}$ square matrix such that ${\bf E}^T{\bf E} = {\bf D}^T{\bf D}$
and
\begin{equation}
\label{eq:def_Y}
  {\bf Y} \triangleq ({\bf D}^{-1})^T{\bf E}^T{\bf X}.
\end{equation}

Since we usually have $m\ll n_\lambda$,
then the formulation \eqref{eq:fcls_reduced_size} opens the door to faster solvers.
Given that ${\bf E}^T{\bf E}$ is positive definite, the equation ${\bf E}^T{\bf E} = {\bf D}^T{\bf D}$
has non-singular solutions. In this paper, we use the Cholesky decomposition to find a
solution of that equation. Note that we have also used solutions based on the eigendecomposition of
${\bf E}^T{\bf E}$, leading to very similar results.

Defining $\bfU \triangleq \bfD \bfA $
and $\bfb^T\triangleq\bs{1}_m^T \bfD^{-1}$, 
the problem \eqref{eq:fcls_reduced_size} can be transformed as
\begin{align}
\begin{aligned}
&\min\limits_{\bfU} \|\bfY - \bfU\|_F^2\\
&\textrm{s.t.} \quad \bfD^{-1}\bfU \geq 0 \quad\textrm{and}\quad \bfb^T\bfU=\bs{1}_n^T.
\end{aligned}
\label{eq:MAP_norm_U}
\end{align}

Obviously, the optimization \eqref{eq:MAP_norm_U} with respect to (w.r.t.)
$\bfU$ can be implemented in parallel for each spectral vector $\bfu_j$, where
$\bfU=[\bfu_1,\cdots,\bfu_n]$ and $\bfu_j$ is the $j$th column of
$\bfU$. In another words, \eqref{eq:MAP_norm_U} can be split into $n$ independent problems
\begin{align}
\begin{aligned}
&\min\limits_{\bfu} \|\bfy_j - \bfu\|_2^2 \\
&\textrm{s.t.} \quad \bfD^{-1}\bfu \geq 0 \quad\textrm{and}\quad \bfb^T\bfu=1
\end{aligned}
\label{eq:opt_ASC_ANC_single}
\end{align}
where $\bfy_j$ is the $j$th column of $\bfY$ $\left(\forall j = 1,\cdots,n \right)$.

Recall now that the Euclidean projection of a given vector $\bfv$ onto a
closed and convex set $\calC$ is defined as \cite{Boyd2004convex} 
\begin{equation}
\Pi_{\calC}(\bfu)\triangleq\argmin_\bfu \left(\|\bfv-\bfu\|_2^2 + \iota_{\calC}(\bfu)\right)
\end{equation}
where $\iota_{\calC}(\bfu)$ denotes the indicator function
\begin{equation}
\iota_{\calC}(\bfu)=
\left\{
\begin{array}{ll}
0 & \textrm{if } \bfu \in \calC \\
\infty & \textrm{otherwise.} \end{array} \right.
\end{equation}
\noindent Therefore, the solution $\hat{\bfu}_j$ of \eqref{eq:opt_ASC_ANC_single} is
the projection of $\bfy_j$ onto the intersection of convex sets
$\calN=\left\{\bfu \in \mathbb{R}^m\,:\bfD^{-1}\bfu\geq 0\right\}$ ({associated with the initial} ANC) and
$\calS=\left\{\bfu \in \mathbb{R}^m\,:\bfb^T\bfu=1\right\}$ ({associated with the initial} ASC)
as follows
\begin{equation}
\label{eq:def_U}
\begin{array}{ll}
\hat{\bfu}_j&=\argmin\limits_{\bfu} \|\bfy_j- \bfu\|_F^2+\iota_{\calN \cap \calS}(\bfu)\\
&=\Pi_{\calN \cap \calS}(\bfy_j)\\
\end{array}
\end{equation}
where $\hat{\bfu}_j$ is the $j$th column of matrix $\hat{\bfU}$.

\begin{remark*}
It is interesting to note that ${\bf Y}$ defined by \eqref{eq:def_Y} can also be  written as $\bfY= \bfD \bfA_{\mathrm{LS}} $ where $\bfA_{\mathrm{LS}} \triangleq \left(\bfE^T\bfE\right)^{-1}\bfE^T\bfX$ is the LS estimator associated with the unconstrained counterpart of \eqref{eq:fcld}. Therefore, ${\bfY}$, $\hat\bfU$ and $\calN \cap \calS$ correspond to
$\bfX$, $\bfA$ and $\calA$, respectively, under the linear mapping induced by $\bfD$.
\end{remark*}

To summarize, supervised SU can be conducted following Algorithm \ref{algo:fun} by first transforming the observation matrix as ${\bf Y} = ({\bf D}^{-1})^T{\bf E}^T{\bf X}$, and then looking for the projection $\hat\bfU$ of $\bfY$ onto $\calN \cap \calS$. Finally, the abundance matrix is easily recovered through the inverse linear mapping $\hat\bfA = \bfD^{-1}\hat{\bfU}$. The projection onto $\calN \cap \calS$ is detailed in the next paragraph.

\begin{algorithm}[h!]
\label{algo:fun}
\KwIn{$\bfX$ (measurements), $\bfE$ (endmember matrix), $\calN$, $\calS$} 
 \tcpp{Calculate the subspace transformation $\bfD$ from the Cholesky decomposition $\bfE^T \bfE= \bfD^T \bfD$} %
 $\bfD \leftarrow$ Chol$\left(\bfE^T \bfE\right)$\; 
 \tcpp{Compute $\bfY$}
 $\bfY \leftarrow  \bfD^{-T} \bfE^T \bfX$\; 
 \tcpp{Project $\bfY$ onto $\calN \cap \calS$ (Algo. \ref{Algo:DAP})}
 $\hat{\bfU} \leftarrow \Pi_{\calN \cap \calS}(\bfY)$\;
 \tcpp{Calculate the abundance}
 $\hat{\bfA} \leftarrow \bfD^{-1}\hat{\bfU}$\;
 \KwOut{$\hat{\bfA}$ (abundance matrix)}
\caption{Fast Unmixing Algorithm}
\DecMargin{1em}
\end{algorithm}

\subsection{Dykstra's Projection onto $\calN\cap\calS$}

While {the matrix $\bfY$} can be computed easily and efficiently from \eqref{eq:def_Y}, {its projection onto $\calN \cap \calS$ following \eqref{eq:def_U} is not
easy to perform}. The difficulty mainly comes from the spectral correlation induced by the linear mapping $\bfD$ in the {non-negativity constraints defining $\calN$}, {which prevents any use of fast algorithms similar to those introduced in \cite{Duchi2008,Kyrillidis2013,Condat2014} dedicated to the projection onto the canonical simplex}. {However, as this set can be regarded as $m$ inequalities, $ \calS \cap\calN$ can be rewritten
as the intersection of $m$ sets
\begin{equation*}
  \calS \cap \calN = \bigcap_{i=1}^m  \calS \cap \calN_i
\end{equation*}
by} splitting $\calN$ into $\calN = \calN_1 \cap \cdots \cap  \calN_m $,
where $\calN_i=\left\{\bfu \in \mathbb{R}^m \,: \bfd_i^T\bfu \geq 0\right\}$ and ${\bf d}_i^T$ represent the $i$th row of $\bfD^{-1}$, i.e.,
$\bfD^{-1}=\left[\bfd_1,\cdots,\bfd_m\right]^T$.
Even though projecting onto this $m$-intersection is difficult, projecting onto each
convex set $ \calS \cap \calN_i$ ($i=1,\ldots,m$) is easier, as it will be shown in paragraph \ref{subsec:projection_individual}. Based on this remark, we propose to perform the projection
onto $\calS \cap \calN$ using the Dykstra's alternating projection algorithm, which was first proposed in \cite{Dykstra1983,Boyle1986}
and has been developed to more general optimization problems \cite{Bauschke2008,Combettes2011}.
More specifically, this projection is split into $m$ iterative projections onto each convex set $\calS\cap\calN_i$ ($i=1,\ldots,m$), following the Dykstra's procedure described in Algorithm \ref{Algo:DAP}.

\begin{algorithm}[h!]
\label{Algo:DAP}
\KwIn{$\bfY$, $\bfD$, $K$}
 \tcpp{Compute $\bfb$}
 $\bfb^T \leftarrow \bs{1}_m^T \bfD^{-1}$\;
 \tcpp{Initialization}
 Set $\bfU_m^{(0)}\leftarrow\bfY$, $\bfQ_1^{(0)}=\cdots=\bfQ_{m}^{(0)}\leftarrow\bf0$\;
 \tcpp{Main iterations}
 \For{$k=1,\cdots,K$}
 {
 \tcpp{Projection onto $\calS \cap \calN_1$ (Algo. \ref{Algo:Proj_NS})}
  $\bfU_1^{(k)} \leftarrow \Pi_{\calS \cap \calN_1}(\bfU_{m}^{(k-1)}+\bfQ_{m}^{(k-1)})$\;
  $\bfQ_{m}^{(k)}\leftarrow\bfU_{m}^{(k-1)}+\bfQ_{m}^{(k-1)}-\bfU_{1}^{(k)}$\;
 \For{$i=2,\cdots,m$}
 {
   \tcpp{Projection onto $\calS \cap \calN_i$ (Algo. \ref{Algo:Proj_NS})}
   $\bfU_i^{(k)} \leftarrow \Pi_{\calS \cap \calN_i}(\bfU_{i-1}^{(k)}+\bfQ_{i-1}^{(k-1)})$\;
  $\bfQ_{i-1}^{(k)}\leftarrow\bfU_{i-1}^{(k)}+\bfQ_{i-1}^{(k-1)}-\bfU_i^{(k)}$\;
 }
 }
 $\hat{\bfU} \leftarrow \bfU_m^{(K)}$\;
 \KwOut{$\hat{\bfU}\leftarrow\Pi_{\calS \cap \calN}\left(\bfY\right)$}
\caption{Dykstra's Projection of $\bfY$ onto $\calS \cap \calN$}
\DecMargin{1em}
\end{algorithm}

The motivations for projecting onto  $\calS \cap \calN_i$ are two-fold.
First, this projection guarantees that the vectors $\hat{\bfu}_j$ always
satisfy {the sum-to-one constraint $\bfb^T \hat{\bfu}_j =1$,
which implies that these vectors never jump out from the hyperplane $\calS$},
and thus accelerates the convergence significantly.
Second, as illustrated later, incorporating the constraint $\bfb^T\bfu=1$ does not increase the projection computational complexity,
which means that projecting onto $\calS \cap \calN_i$ is as easy as projecting onto $\calN_i$
(for $i=1,\cdots,m$). The projection onto $\calS \cap \calN_i$ is described in the next paragraph.


\subsection{Projection onto $\calS \cap \calN_i$}
\label{subsec:projection_individual}
The main step of the Dykstra's alternating procedure (Algorithm \ref{Algo:DAP}) consists of computing the projection
$\bfU^\ast_i$ of a given matrix $\bfZ$ onto the set $\calS \cap \calN_i$
\begin{align*}
\bfU^\ast_i & = \Pi_{\calS \cap \calN_i}\left(\bfZ\right) \\
            & \equiv [\Pi_{\calS \cap \calN_i}(\bfz_1),\dots, \Pi_{\calS \cap \calN_i}(\bfz_n)].
\end{align*}
Let ${\bf z}\in \mathbb{R}^{m}$ denote a generic column of $\bf Z$. The computation
of the projection $\Pi_{\calS \cap \calN_i}(\bfz)$
 can be achieved by solving the
following convex constrained optimization problem:
\begin{equation}
\begin{split}
&\min\limits_{\bfu} \|\bfz - \bfu\|_2^2 \\
&\textrm{s.t.} \quad \bfd_i^T \bfu\geq0 \quad \textrm{and} \quad \bfb^T\bfu=1.
\end{split}
\label{eq:opt_ASC_ANC_a}
\end{equation}

\begin{figure}
\centering
\includegraphics[width = 8cm]{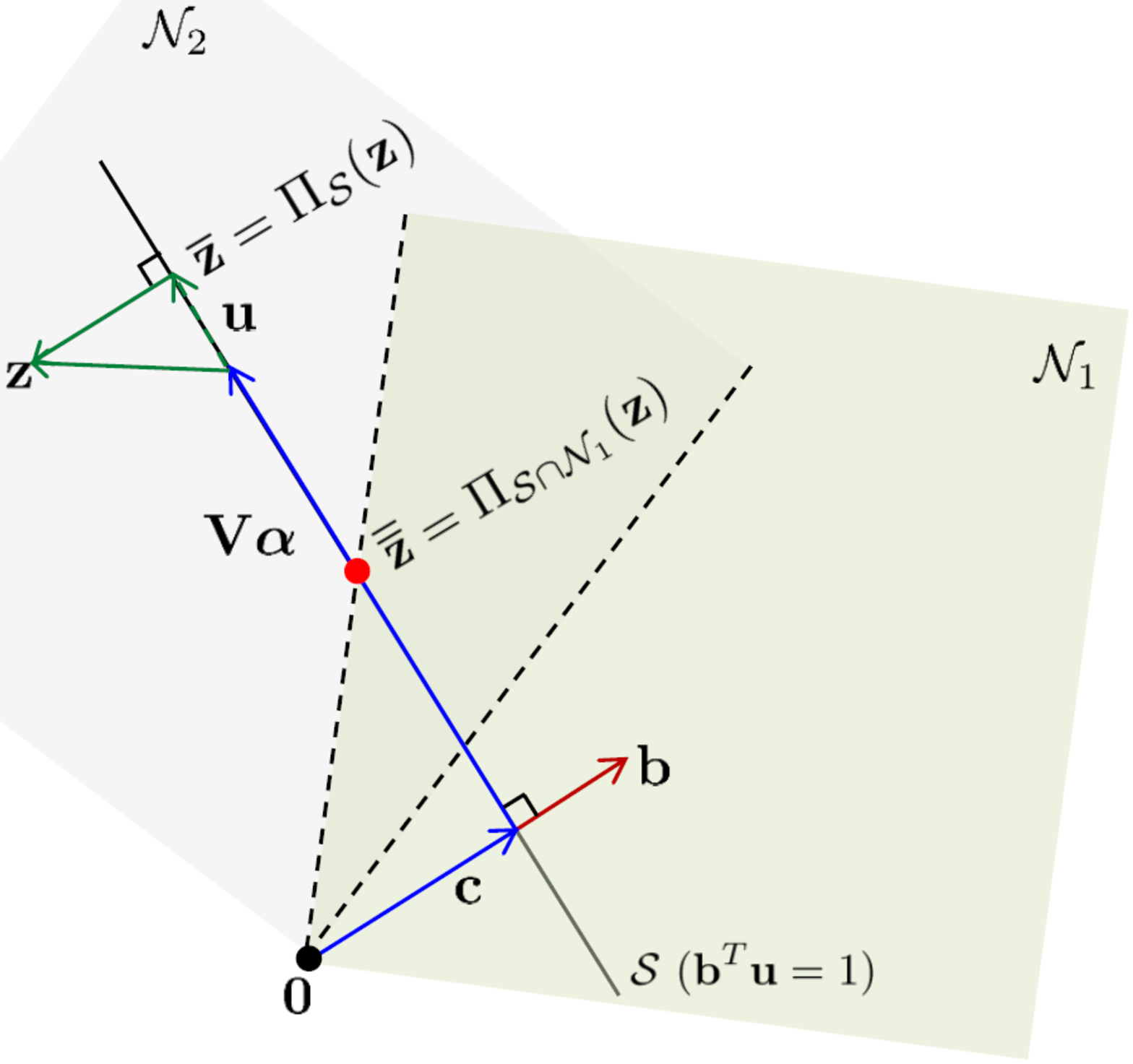} 
\caption{Illustration of the projection of $\bf z$ onto the set ${\cal S}\cap {\cal N}_1$: the  set $\cal S$ is defined by  the vector ${\bf c}\in\cal S$ and by the vector $\bf b$ orthogonal to the subspace ${\cal S}-\{\bf c\}$; the vector ${\bf u}\in \cal S$ may be written as ${\bf u = V}\bs{\alpha}+{\bf c}$ where $\bf V$ spans the subspace ${\cal S}-\{\bf c\}$ and $\bs{\alpha} \in\mathbb{R}^{(m-1)}$; the vector $\overline{\bf z}$ is the orthogonal projection of $\bf z$ onto $\cal S$; the vector $\overline{\overline{\bf z}}$ is the orthogonal projection of $\overline{\bf z}$ onto ${\cal S}\cap {\cal N}_1$, which is also the orthogonal projection of $\bf z$ onto the set ${\cal S}\cap {\cal N}_1$.
}
\label{fig:affine_half_planes}
\vspace{-0.5cm}
\end{figure}

{To solve the optimization (\ref{eq:opt_ASC_ANC_a}), we start by removing the
constraint $\bfb^T\bfu=1$ by an appropriate change of variables.
Having in mind that  the set ${\cal S} =\{ {\bf u}\in\mathbb{R}^m:{\bf b}^T{\bf u} = 1\,\}$  is a hyperplane} that contains the vector ${\bf c} = {\bf b}/\|{\bf b}\|^2_2$, then that constraint is equivalent to ${\bfu = {\bfc} }+{\bfV}\bs{\alpha}$, where $\bs{\alpha} \in \mathbb{R}^{m-1}$ and the columns of ${\bf V} \in\mathbb{R}^{m\times(m-1)}$ span the subspace ${\cal S}-\{\bfc\} =\{ {\bfu}\in \mathbb{R}^m\,:{\bfb}^T{\bfu} = 0\,\}$, of dimension $(m-1)$. The matrix  ${\bf V}$ is chosen such that ${\bf V}^T{\bf V}={\bf I}_{m-1}$, {\em i.e.}, the columns of $\bf V$ are orthonormal.  Fig. \ref{fig:affine_half_planes} schematizes the mentioned entities jointly with $\overline{\bf z}$, the orthogonal projection of $\bf z$ onto $\cal S$,  and 
$\overline{\overline{\bf z}}$,  the orthogonal projection of $\overline{\bf z}$ onto ${\cal S}_1\cap {\cal N}_1$. The former projection  may be written as
\begin{align}
   \nonumber
   \overline{\bf z} & \equiv \Pi_{\cal S}({\bf z})\\
     \label{eq:proj_S}
                & = {\bf c} + {\bf P}({\bf z -c})
\end{align}
where ${\bf P} \equiv {\bf VV}^T = {\bfI}_m-{\bf bb}^T/\|{\bf b}\|_2^2$
denotes the orthogonal projection matrix onto ${\cal S}-\{\bf c\}$. {With these objects in place,  and given ${\bf z}\in\mathbb{R}^m$ and $\bf u\in {\cal S}$, we  simplify the cost function $\|{\bf z-u}\|^2_2$ by introducing the projection of $\bf z$ onto $\cal S$ and by using the Pythagorean theorem as follows:}
\begin{align}
  \nonumber
  \|{\bf z -u}\|_2^2 & = \|{\bf z}- \overline{\bf z}\|_2^2 + 
 \|\overline{\bf z}-{\bf u}\|_2^2\\
   \nonumber
             & = \|{\bf z}- \overline{\bf z}\|_2^2 + 
 \|(\overline{\bf z}-{\bf c})-{\bf V}\bs{\alpha}\|_2^2\\
 \label{eq:proj_error_c} 
              & = \|{\bf z}- \overline{\bf z}\|_2^2 + 
 \|{\bf V}^T({\bf z-c})-\bs{\alpha}\|_2^2
\end{align}
where the right hand term in (\ref{eq:proj_error_c}) derives directly from  (\ref{eq:proj_S}) and from the fact that ${\bf V}^T{\bf V}={\bf I}_{m-1}$. 
By introducing ${\bf u = c+V}\bs{\alpha}$ in \eqref{eq:opt_ASC_ANC_a}, we obtain the  equivalent optimization
\begin{align}
\min\limits_{\bs{\alpha}} \|{\bf V}^T({\bf z-c}) - \bs{\alpha}\|_2^2
\;\;\textrm{s.t.} \;\; ({\bf V}^T{\bf d}_i)^T\bs{\alpha}\geq  -({\bf d}_i^T{\bf c})
\label{eq:opt_ASC_ANC_b}
\end{align}
which is a projection onto a half space whose solution is  \cite{Boyd2004convex}
\[\bs{\alpha}^* = {\bf V}^T({\bf z-c})  + \tau_{i}\frac{{\bf V}^T{\bf d}_i}{\|{\bf V}^T{\bf d}_i\|_2}\]
where 
\begin{align*}
\tau_{i} &= \max\left\{0, -\frac{{\bf d}_{i}^T {\bf V}}{\|{\bf V}^T{\bf d}_i\|_2}\big({\bf V}^T({\bf z -c})\big)- \frac{{\bf d}_i^T{\bf c}}{\|{\bf V}^T{\bf d}_i\|_2}\right\}\\
      & = \max\{0, -{\bf s}_i^T{\bf z}+f_i\}
\end{align*}
with  ${\bf s}_i \equiv {\bf P d}_i/\|{\bf P d}_i\|_2$, $f_i \equiv -{\bf d}_i^T{\bf c}/\|{\bf P}{\bf d}_i\|_2$, and we have used the facts that $\|{\bf V}^T{\bf x}\|_2=\|{\bf P}{\bf x}\|_2$ and ${\bf V}^T{\bf c} = \bf 0$.

Recalling that ${\bf u = {\bf c} }+{\bf V}\bs{\alpha}$, we obtain
\begin{equation}
\label{eq:proj_SN}
\begin{split}
\overline{\overline{\bf z}} & = {\bf c} + {\bf VV}^T({\bf z-c})  + \tau_{i}{\bf s}_i\\
                  & = \Pi_{\cal S}({\bf z}) + \tau_{i}{\bf s}_{i}.
\end{split}
\end{equation}
The interpretation of (\ref{eq:proj_SN}) is clear: the orthogonal projection of $\bf z$ onto ${\cal S}\cap {\cal N}_i$ is obtained by  first computing $\overline{\bf z} = \Pi_{\cal S}({\bf z})$, {\em i.e.}. the  projection ${\bf z}$ onto the hyperplane $\cal S$, and then computing $\overline{\overline{\bf z}} = \Pi_{{\cal S}\cap {\cal N}_i}(\overline{\bf z})$, {\em i.e.}. the  projection $\overline{\bf z}$ onto the  intersection  ${\cal S}\cap {\cal N}_i$.  Given that ${\cal S}\cap {\cal N}_i \subset {\cal S}$, then (\ref{eq:proj_SN}) is, essentially, a consequence of a well know result: given  a convex set  contained in some subspace, then the orthogonal projection of any point in the convex set can be accomplished by first projecting orthogonally on that subspace, and then projecting the result on the convex set \cite[Ch. 5.14]{Deutsch2001}.  

Finally, computing ${\bf U}^*_{i}$  can be conducted in parallel for each column of $\bfZ$ leading to the following matrix update rule summarized in Algorithm \ref{Algo:Proj_NS}):
\begin{align}
\label{eq:Proj_Geometry}
   {\bf U}^*_{i} & = \Pi_{\cal S}({\bf Z}) + {\bf s}_{i}\bs{\tau}_{i}^T
\end{align}
with $\bs{\tau}_i^T \in\mathbb{R}^{1\times n}$ given by
\[
 \bs{\tau}_i^T = \max\{{\bf 0}, f_{i}{\bf 1}_n^T-  {\bf s}_{i}^T{\bf Z}\}
\]
where
\begin{align}
f_{i} & = -\frac{{\bf d}_i^T{\bf c}}{\|{\bf P d}_i\|_2} 
\end{align}
and the operator $\max$ has to be understood in the component-wise sense

\begin{algorithm}[h!]
\label{Algo:Proj_NS}
\KwIn{$\bfZ$, $\bfb$, $\bfd_i$}
  \tcpp{Calculate $\bfP \bfd_i$, $\bfs_i$, $\bfc$ and $f_i$}
  $\bfc \leftarrow \bfb/\|{\bf b}\|^2_2$\;
  $\bfP \bfd_i \leftarrow \bfd_i- {\bf cb}^T \bfd_i$\;
  $\bfs_i \leftarrow \bfP \bfd_i / \|\bfP \bfd_i\|_2$ \;
  $f_{i}  \leftarrow -{{\bf d}_i^T{\bf c}}/{\|\bfP \bfd_i\|_2}$ \;
  \tcpp{Calculate $\bs{\tau}_i^T$}
  $\bs{\tau}_i^T \leftarrow \max\{{\bf 0}, f_{i}{\bf 1}_n^T-  {\bf s}_{i}^T{\bf Z}\}$\;

  \tcpp{Project $\bfZ$ onto $\calS$}
  $\Pi_{\cal S}({\bf Z})\leftarrow{\bf c}{\bf 1}_n^T + {\bf P}(\bfZ - {\bf c}{\bf 1}_n^T)$\;
  \tcpp{Compute the final solution $\bfU_i^\ast$}
  $\bfU_i^\ast \leftarrow\Pi_{\cal S}({\bf Z}) + {\bf s}_{i}\bs{\tau}_{i}^T$\;
 \KwOut{$\bfU_i^\ast$}
\caption{Projecting $\bfZ$ onto $\calS \cap \calN_i$}
\DecMargin{1em}
\end{algorithm}

Note that using the Karush-Kuhn-Tucker (KKT) conditions to solve the problem  \eqref{eq:opt_ASC_ANC_a}
can also lead to this exact solution, as described in the Appendix.

\subsection{Convergence Analysis}
\label{subsec:conv_als}
The convergence of the Dykstra's projection was first proved in \cite{Dykstra1983},
where it was claimed that the sequences generated using Dykstra's algorithm are guaranteed to
converge to the projection of the original point onto the intersection of the convex sets.
Its convergence rate was explored later \cite{Deutsch1994rate,Xu2000}.
We now recall the Deutsch-Hundal theorem providing the convergence rate
of the projection onto the intersection of $m$ closed half-spaces.

\begin{theorem}[Deutsch-Hundal, \cite{Deutsch1994rate}; Theorem 3.8]
\label{Theo:Conv_Rate}
Assuming that $\bfX_k$ is the $k$th projected result in Dykstra's algorithm
and $\bfX_{\infty}$ is the converged point,
there exist constants $0\leq c <1$ and $\rho >0$ such that
\begin{equation}
\|\bfX_k-\bfX_{\infty}\|_F^2 \leq \rho c^k
\end{equation}
for all $k$.
\end{theorem}
Theorem \ref{Theo:Conv_Rate} demonstrates that Dykstra's projection has a
linear convergence rate \cite{Wright1999}. The convergence speed depends on
the constant $c$, which depends on the number of constraints $m$ and the `angle' between two
half-spaces \cite{Deutsch1994rate}. To the best of our knowledge, the explicit form
of $c$ only exists for $m=2$ half-spaces and its determination for $m>2$ is still an open
problem \cite{Deutsch2008rate}.

\subsection{Complexity Analysis}
\label{subsec:comp_als}
To summarize, the projection onto $\calS \cap \calN$ can be obtained by iteratively projecting onto the $m$
sets $\calS \cap \calN_i$ ($i=1,\ldots,m$) using a Dykstra's projection scheme as described in Algorithm \ref{Algo:DAP}.
The output of this algorithm converges to the projection of the initial point $\bfY$ onto $\calS \cap \calN$. It is interesting
to note that the quantities denoted as $\Pi_{\cal S}({\bf Z})$ in Algorithm \ref{Algo:Proj_NS} needs to be calculated only once 
since the projection of $\bfZ$ will be itself $\bfZ$ from the second projection $\Pi_{\calS \cap \calN_2}$. This results from the fact that
the projection never jumps out from the hyperplane $\calS$.

Moreover, the most computationally expensive part of the proposed unmixing algorithm (Algorithm \ref{algo:fun}) 
is the iterative procedure to project onto $\calS \cap \calN$, as described in Algorithm \ref{Algo:DAP}. For each iteration,
the heaviest step is the projection onto the intersection $\calS \cap \calN_i$ summarized in
Algorithm \ref{Algo:Proj_NS}. With the proposed approach, this projection only requires vector products and
sums, with a cost of $\mathcal{O}(nm)$ operations, contrary to the $\mathcal{O}(nm^2)$ computational cost of \cite{Heylen2013}.
Thus, each iteration of Algorithm \ref{Algo:DAP} has a complexity of order $\mathcal{O}(nm^2)$. 

%% file: simulation.tex
This section compares the proposed unmixing algorithm
with several state-of-the-art unmixing algorithms, i.e.,
FCLS \cite{Heinz2001}, SUNSAL \cite{Bioucas2010SUNSAL}, IPLS \cite{Chouzenoux2014}
and APU \cite{Heylen2013}. All algorithms have been implemented using MATLAB R2014A on a computer with
Intel(R) Core(TM) i7-2600 CPU@3.40GHz and 8GB RAM. To conduct a fair comparison, they have been implemented
in the signal subspace without using any parallelization. These unmixing algorithms have been compared using 
the figures of merit described in Section \ref{subsec:performance}. Several experiments have been conducted using
synthetic datasets and are presented in Section \ref{subsec:synthetic}. Two real hyperspectral (HS) datasets 
associated with two different applications are considered in Section \ref{subsec:real}. The MATLAB codes
and all the simulation results are available on the first author's homepage\footnote{\url{http://wei.perso.enseeiht.fr/}}. 

\subsection{Performance Measures}
\label{subsec:performance}
In what follows, $\widehat{\bf A}_t$
denotes the estimation of $\bfA$ obtained at time $t$ (in seconds) for a given algorithm. Provided that the endmember matrix $\bf E$ has full column rank, the solution of \eqref{eq:fcld}
is unique and all the algorithms are expected to converge to this unique solution, denoted
as $\bfA^{\star} \triangleq \widehat{\bf A}_\infty$ (ignoring numerical errors). In this work, one of the state-of-the-art methods is run with a large number of iterations
($n=5000$ in our experiments) to guarantee that the optimal point $\bfA^{\star}$ has been reached. 


\subsubsection{Convergence Assessment}
First, different solvers designed to compute the solution
of \eqref{eq:fcld}  have been compared w.r.t. the time they require to
achieve a given accuracy. Thus, all these algorithms have been run on the same platform and we
have evaluated the relative error (RE) between $\widehat{\bfA}_t$ and $\bfA^{\star}$
as a function of the computational time defined as
\begin{align*}
    \text{RE}_t = \frac{\|\widehat{\bf A}_t- {\bf A}^{\star}\|^2_F}{\| {\bfA}^{\star}\|^2_F}.
\end{align*}

\subsubsection{Quality Assessment}
To analyze the quality of the unmixing results, we have also considered the
normalized mean square error (NMSE) 
\begin{align*}
    \textrm{NMSE}_t = \frac{\|\widehat{\bf A}_t- {\bf A}\|^2_F}{\| {\bf A}\|^2_F}.
\end{align*}
The smaller NMSE$_t$, the better the quality of the unmixing. 
Note that $\text{NMSE}_\infty=\frac{\|{\bfA}^{\star}-\bfA\|^2_F}{\|\bfA\|^2_F}$
is a characteristic of the objective criterion \eqref{eq:fcld} and not of the algorithm.

\subsection{Unmixing Synthetic Data}
\label{subsec:synthetic}
\begin{figure}[h!]
\centering
\includegraphics[width=0.49\columnwidth]{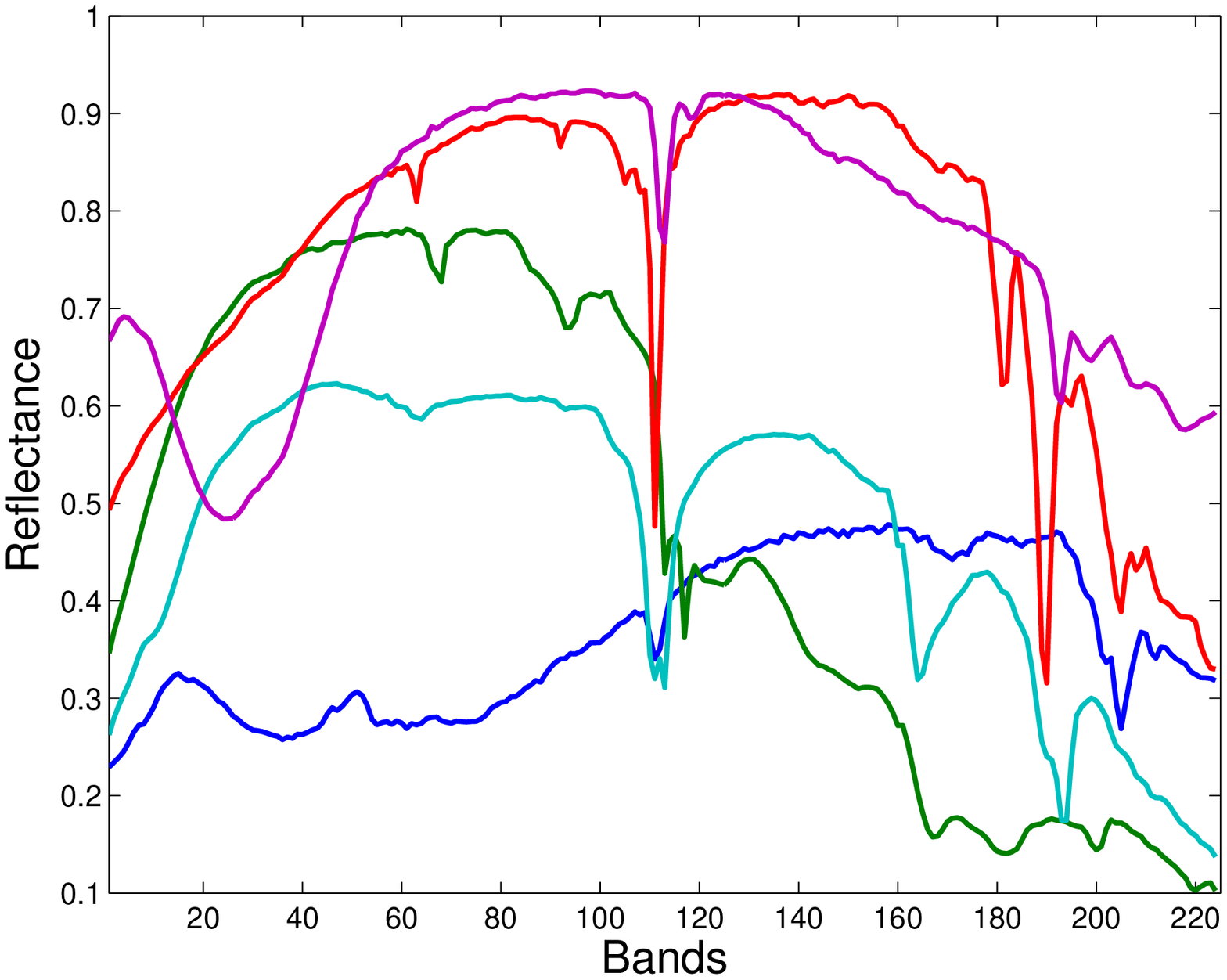}
\includegraphics[width=0.49\columnwidth]{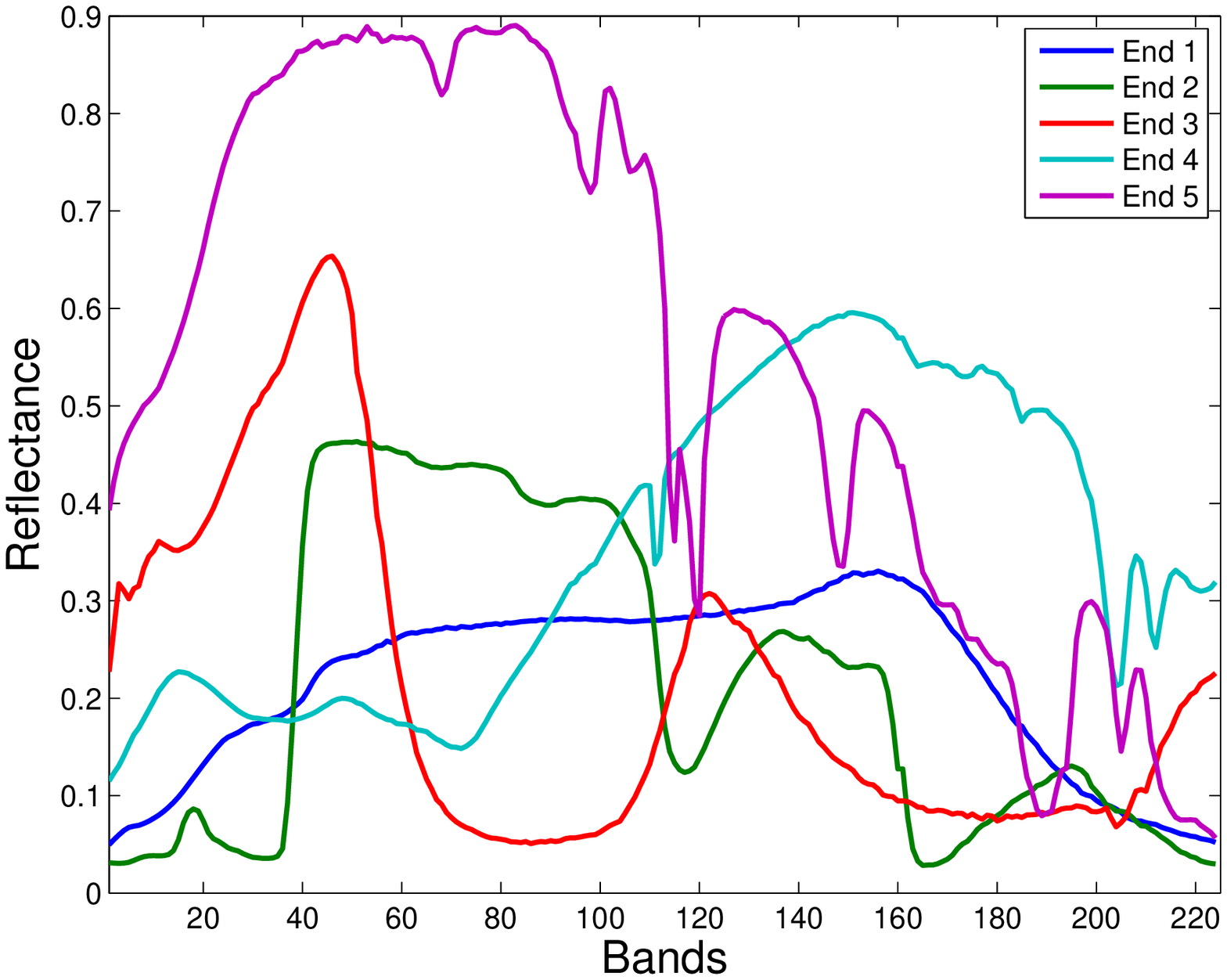}
\caption{Five endmember signatures randomly selected from $\bfE_3$ (left) and $\bfE_{20}$ (right).}
\label{fig:syn_end_sig}
\end{figure}

The synthetic data is generated using endmember spectra selected
from the United States Geological Survey (USGS) digital spectral
library\footnote{http://speclab.cr.usgs.gov/spectral.lib06/}.
These reflectance spectra consists of $L = 224$  spectral bands
from $383$nm to $2508$nm.
To mitigate the impact of the intra-endmember correlation, three different
subsets $\bfE_3$, $\bfE_{10}$ and $\bfE_{20}$ have been built from this USGS library.
More specifically, $\bfE_\alpha$ is an endmember matrix in which the angle between
any two different columns (endmember signatures) is larger than
$\alpha$ (in degree). Thus, the smaller $\alpha$, the more similar the endmembers and 
the higher the conditioning number of $\bfE$.
For example, $\bfE_3$ contains similar endmembers with very small variations (including
scalings) of the same materials and $\bfE_{20}$ contains endmembers
which are relatively less similar. As an illustration, a random selection of several endmembers
from $\bfE_3$ and $\bfE_{20}$ have been depicted in Fig. \ref{fig:syn_end_sig}. 
The abundances have been generated uniformly in the simplex $\calA$
defined by the ANC and ASC constraints. 

Unless indicated, the performance of these algorithms has been
evaluated on a synthetic image of size ${100} \times {100}$ whose signal 
to noise ratio (SNR) has been fixed to SNR=$30$dB and the number of considered endmembers is $m=5$.

\subsubsection{Initialization}
The proposed SUDAP, APU and FCLS algorithms do not require any initialization
contrary to SUNSAL and IPLS. As suggested by the authors of these two methods,
SUNSAL has been initialized with the unconstrained LS
estimator of the abundances 
whereas IPLS has been initialized with the zero matrix. Note that our simulations
have shown that both SUNSAL and IPLS are not sensitive to these
initializations.

\subsubsection{Performance \emph{vs.} Time}
\label{subsubsec:perf_time}
\begin{figure}[h!]
\centering
    \includegraphics[width=0.49\columnwidth]{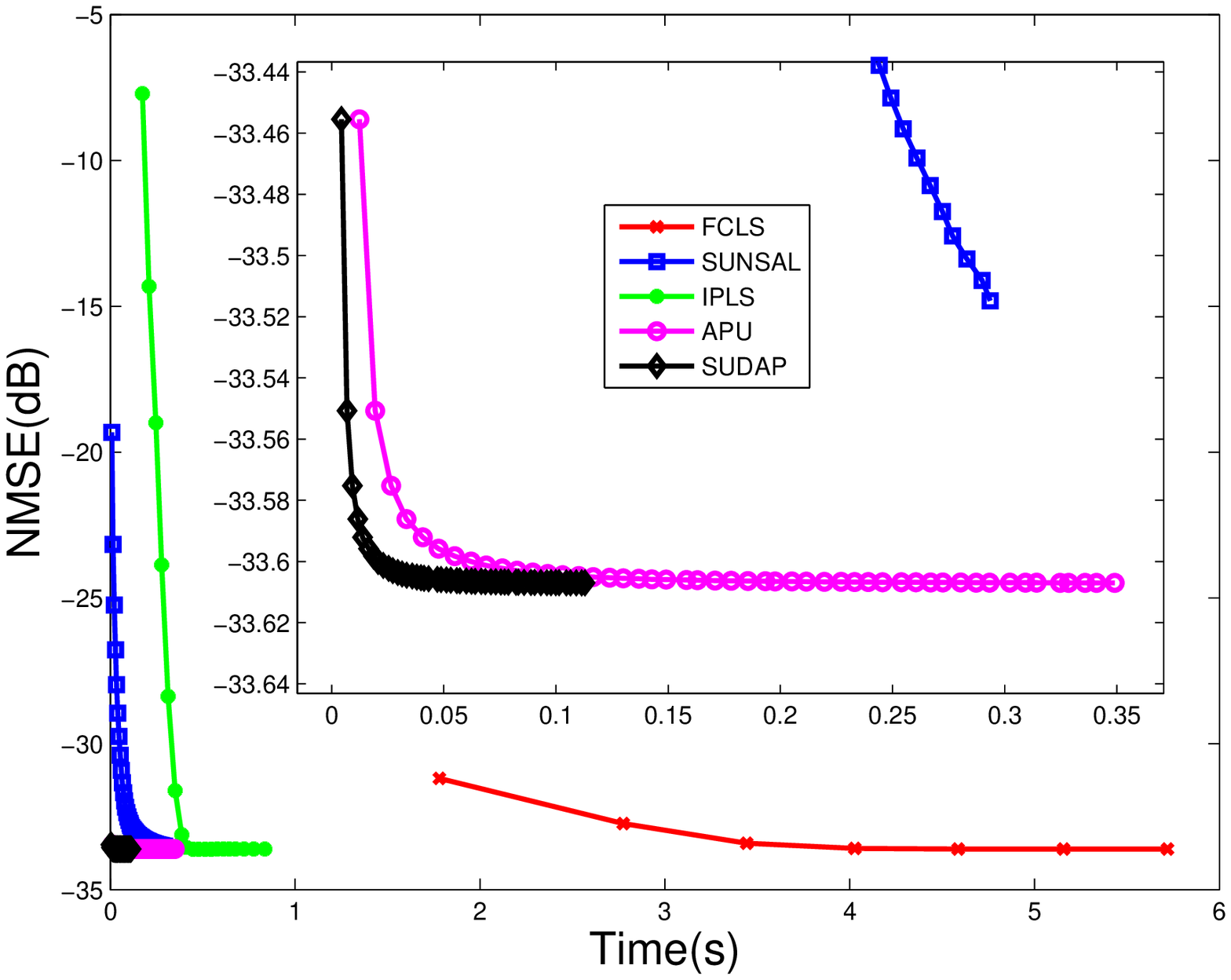}
    \includegraphics[width=0.49\columnwidth]{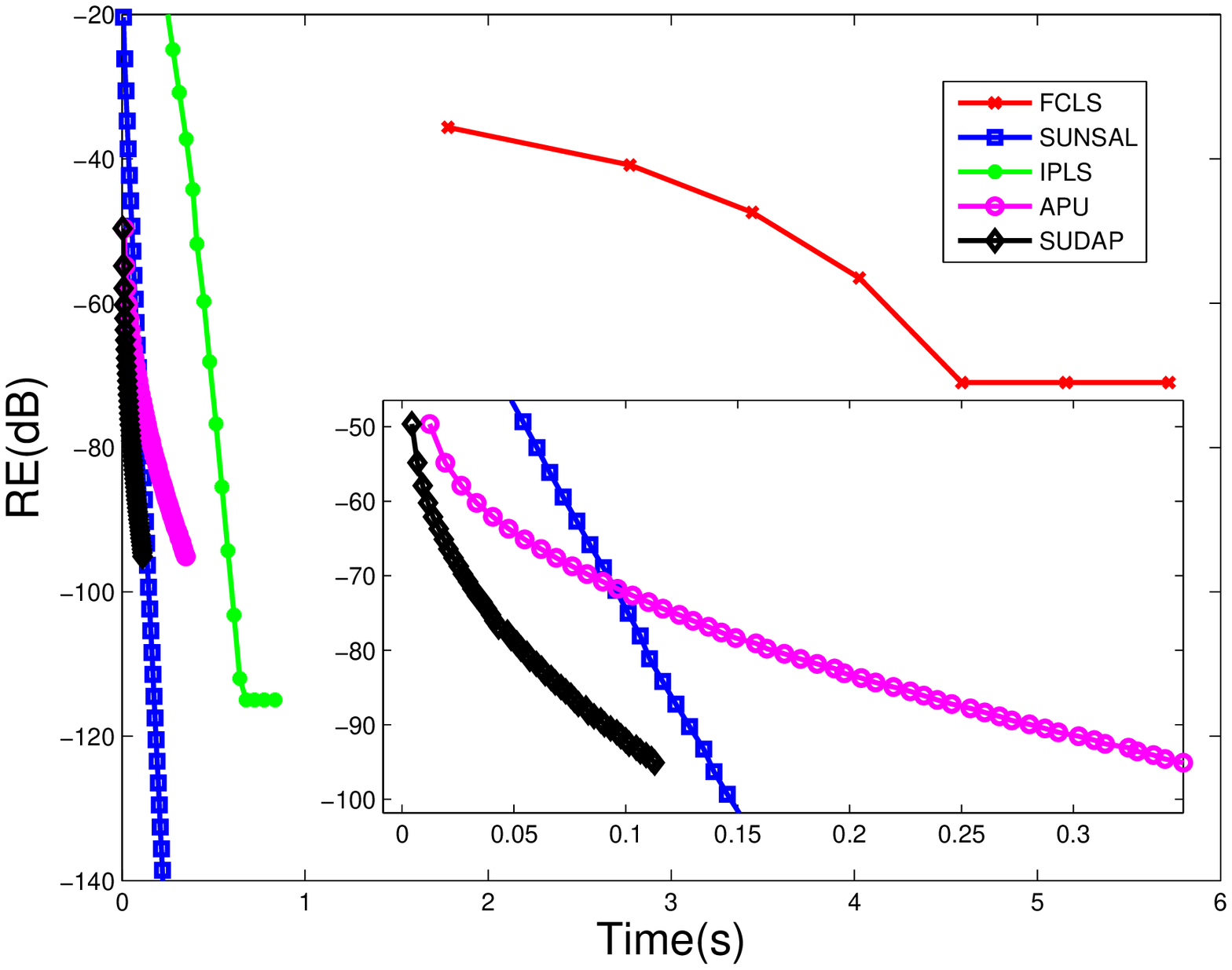}
    \caption{NMSE (left) and RE (right) \emph{vs.} time (zoomed version in top right).}
\label{fig:NMSE_time}
\end{figure}

The NMSE and RE for these five different algorithms
are displayed in Fig. \ref{fig:NMSE_time} 
as a function of the execution time.
These results have been obtained by averaging the outputs of $30$ Monte 
Carlo runs. More precisely, 10 randomly selected matrices for each set 
$\bfE_3$, $\bfE_{10}$ and $\bfE_{20}$ are used to consider the different
intra-endmember correlations.
All the algorithms converge to the same solution as expected. However, as demonstrated in these
two figures, SUNSAL, APU and the proposed SUDAP are much faster than FCLS
and IPLS. From the zoomed version in Fig. \ref{fig:NMSE_time},
we can observe that in the first iterations SUDAP converges faster than
APU and SUNSAL. More specifically, for instance, if the respective algorithms 
are stopped once RE$_t<-80$dB (around $t=50$ms),
SUDAP performs faster than SUNSAL and APU and with a lower NMSE$_t$.


%

\subsubsection{Time \emph{vs.} the Number of Endmembers}
\label{subsubsec:time_end}
\begin{figure}[h!]
\includegraphics[width=0.49\columnwidth]{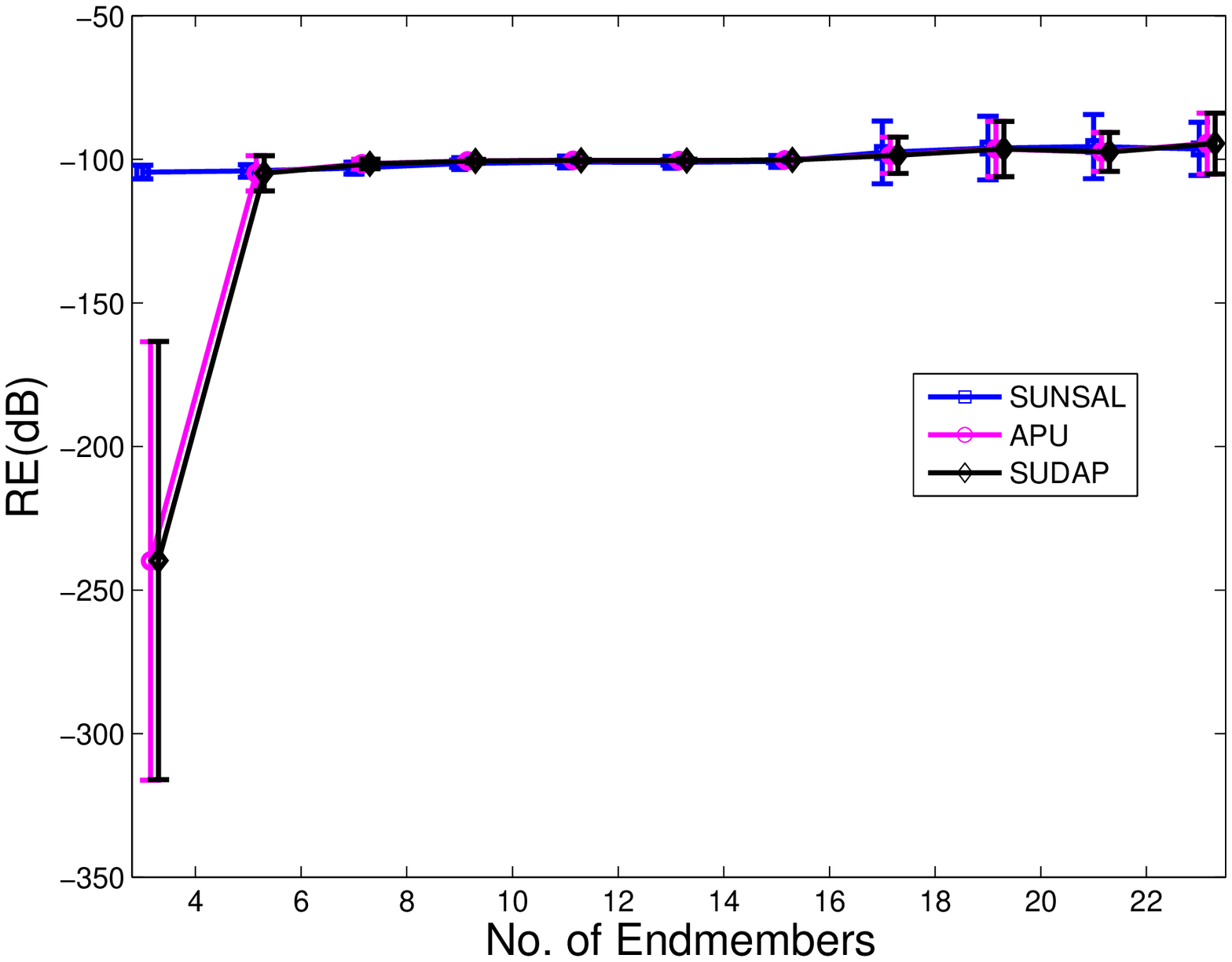}
\includegraphics[width=0.49\columnwidth]{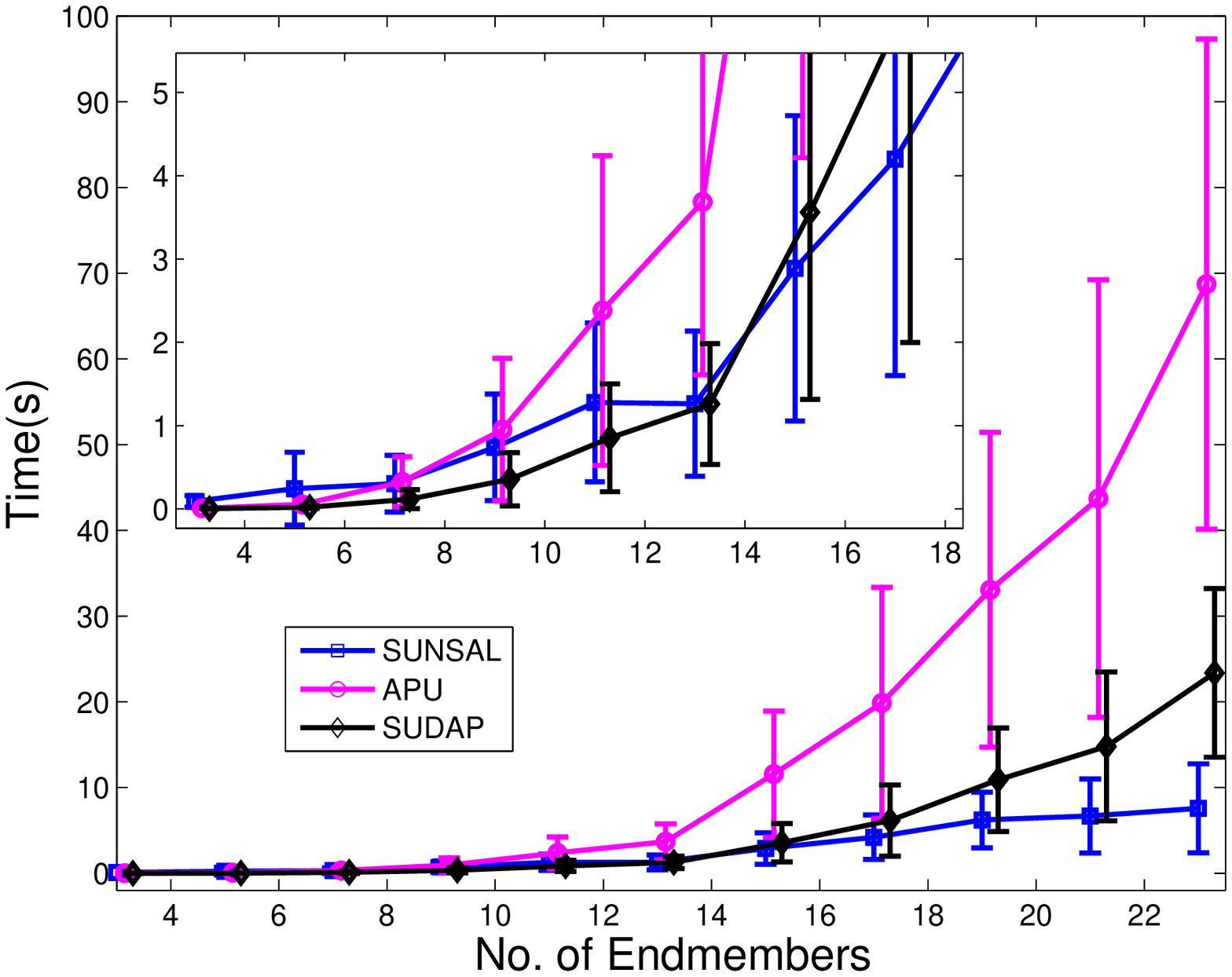}
\caption{RE (left) and time (right) \emph{vs.} number of endmembers for SUNSAL, APU and SUDAP (RE$_t<-100$dB).}
\label{fig:RE_End}
\end{figure}

In this test, the number of endmembers $m$ varies from $3$ to $23$ while the other
parameters have been fixed to the same values as in Section \ref{subsubsec:perf_time} (SNR$=30$dB and $n=100^2$).
The endmember signatures have been selected from $\bfE_{10}$ (similar results have been observed when using
$\bfE_{3}$ and $\bfE_{20}$). All the algorithms have been
stopped once $\hat{\bfA}_t$ reaches the same convergence criterion RE$_t<-100$dB. The proposed SUDAP has been compared with the two most
competitive algorithms SUNSAL and APU. The final REs and the corresponding computational times
versus $m$ have been reported in Fig. \ref{fig:RE_End}, including error bars to monitor the stability 
of the algorithms (these results have been computed from $30$ Monte Carlo runs). 

Fig. \ref{fig:RE_End} (left) shows that all the algorithms have converged to a point
satisfying RE$_t<-100$dB and that SUDAP and APU are slightly better than SUNSAL.
However, SUNSAL provides a smaller estimation variance leading to a more stable estimator.
Fig. \ref{fig:RE_End} (right) shows that the execution time of the three methods is an
increasing function of the number of endmembers $m$, as expected. However, there are significant differences
between the respective rates of increase. The execution times of APU and SUDAP are cubic and
quadratic functions of $m$ whereas SUNSAL benefits from a milder increasing rate.
More precisely, SUDAP is faster than SUNSAL when the number of endmembers is small, e.g., smaller than $19$ (this
value may change depending on the SNR value, the conditioning number of $\bf E$, the abundance
statistics, etc.). Conversely, SUNSAL is faster than SUDAP for $m\geq 19$. SUNSAL is more efficient
than APU for $m \geq 15$ and SUDAP is always faster than APU. The error bars confirm that SUNSAL offers
more stable results than SUDAP and APU. Therefore, it can be concluded that the proposed SUDAP is more 
promising to unmix a multi-band image containing a reasonable number of materials, while SUNSAL is more efficient
when considering a scenario containing a lot of materials.

\subsubsection{Time \emph{vs.} Number of Pixels}
\begin{figure}[h!]
\centering
\includegraphics[width=0.49\columnwidth]{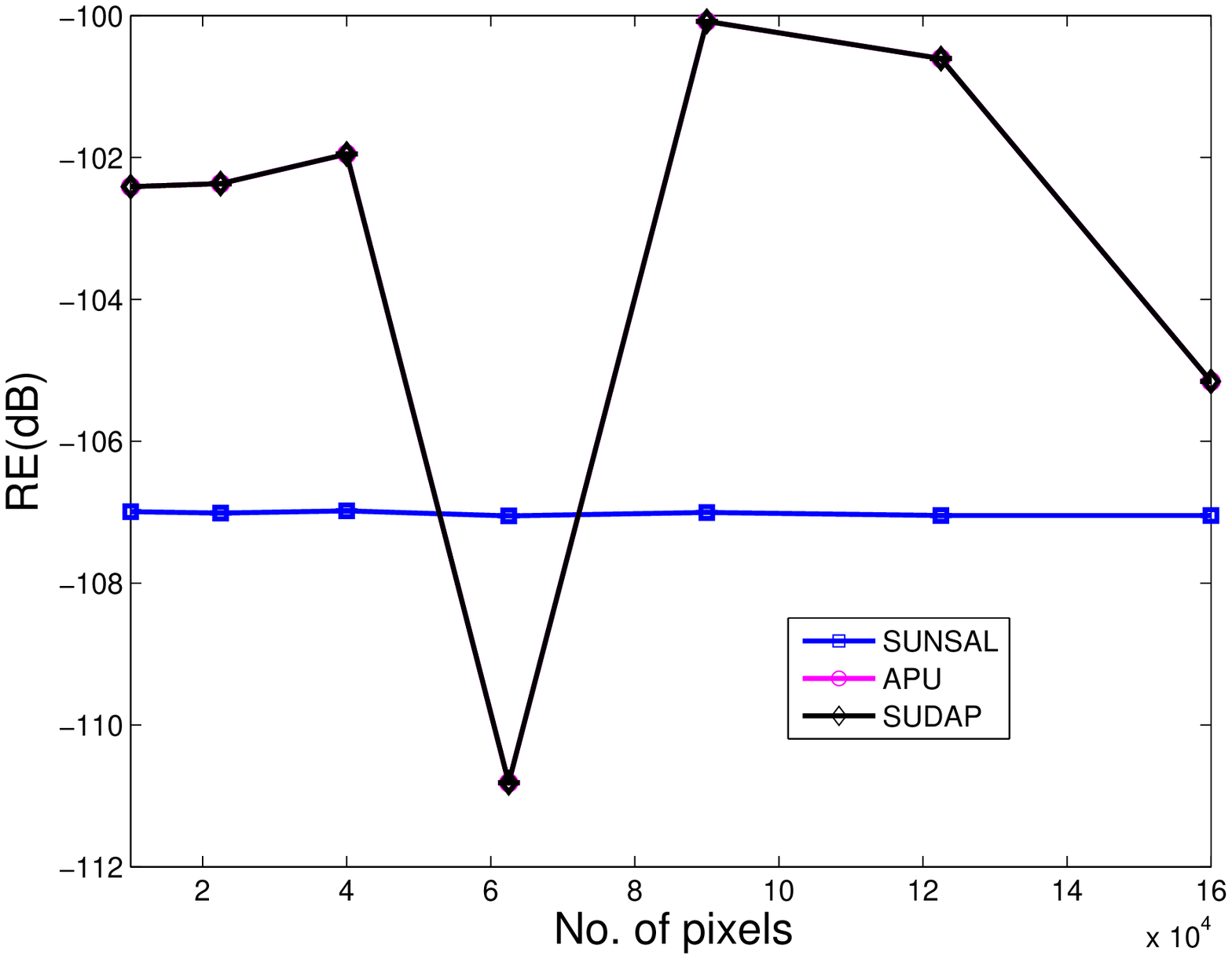}
\includegraphics[width=0.49\columnwidth]{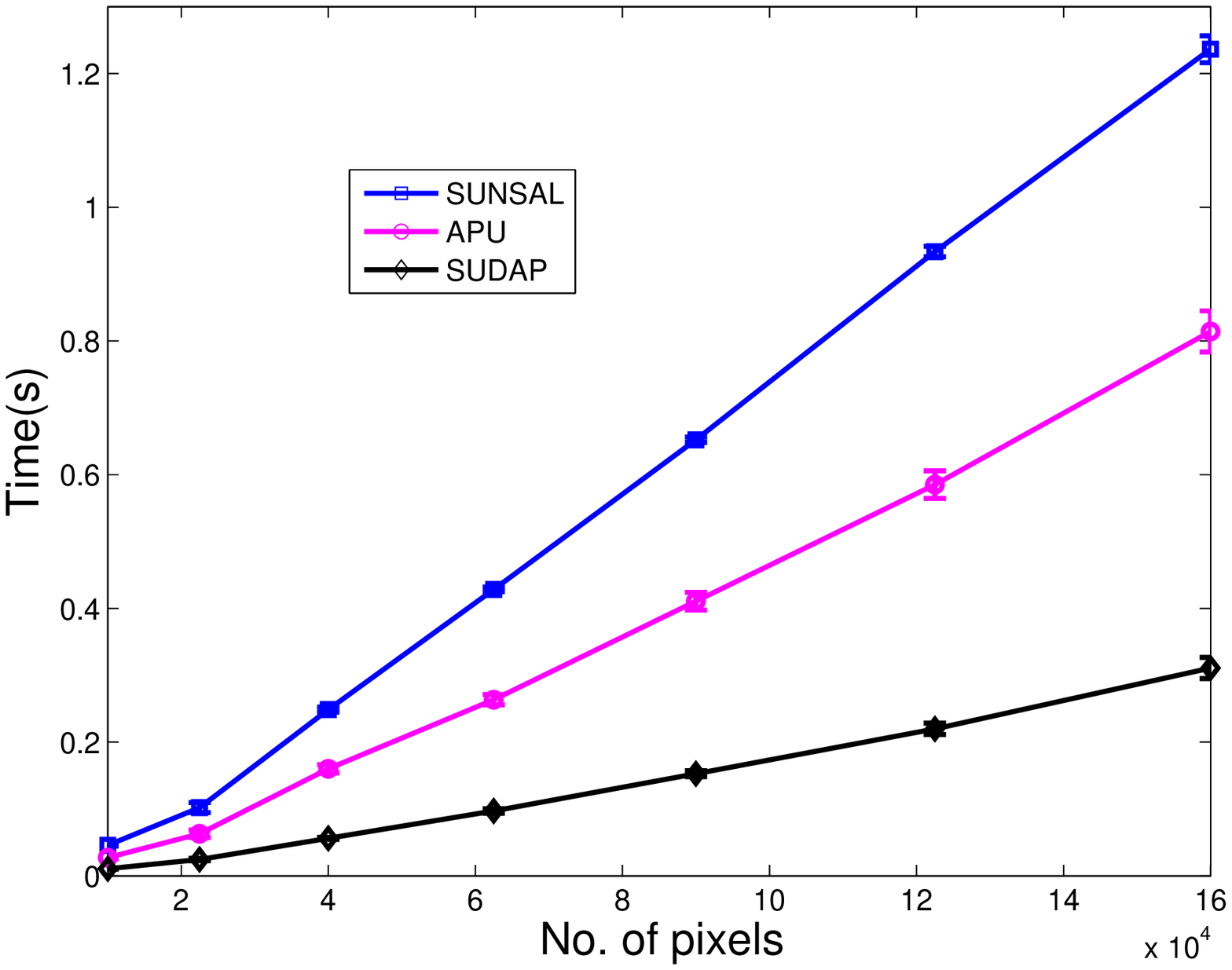}
\caption{RE (left) and time (right) \emph{vs.} number of pixels for SUNSAL, APU and SUDAP (RE$_t<-100$dB).}
\label{fig:RE_pixels}
\end{figure}

In this test, the performance of the algorithms has been evaluated for a varying number of pixels $n$ from $100^2$ to $400^2$ 
(the other parameters have been fixed the same values as in Section \ref{subsubsec:perf_time}). 
The endmember signatures have been selected from $\bfE_{10}$ (similar results have been observed when using
$\bfE_{3}$ and $\bfE_{20}$) and the stopping rule has been chosen as RE$_t< -100$dB. All results have been averaged 
from $30$ Monte Carlo runs. The final REs and the corresponding computational times are shown in Fig. \ref{fig:RE_pixels}.
The computational time of the three algorithms increases approximately linearly w.r.t. the
number of image pixels and SUDAP provides the faster solution, regardless the number of pixels.

\subsubsection{Time \emph{vs.} SNR}
\begin{figure}[h!]
\centering
\includegraphics[width=0.49\columnwidth]{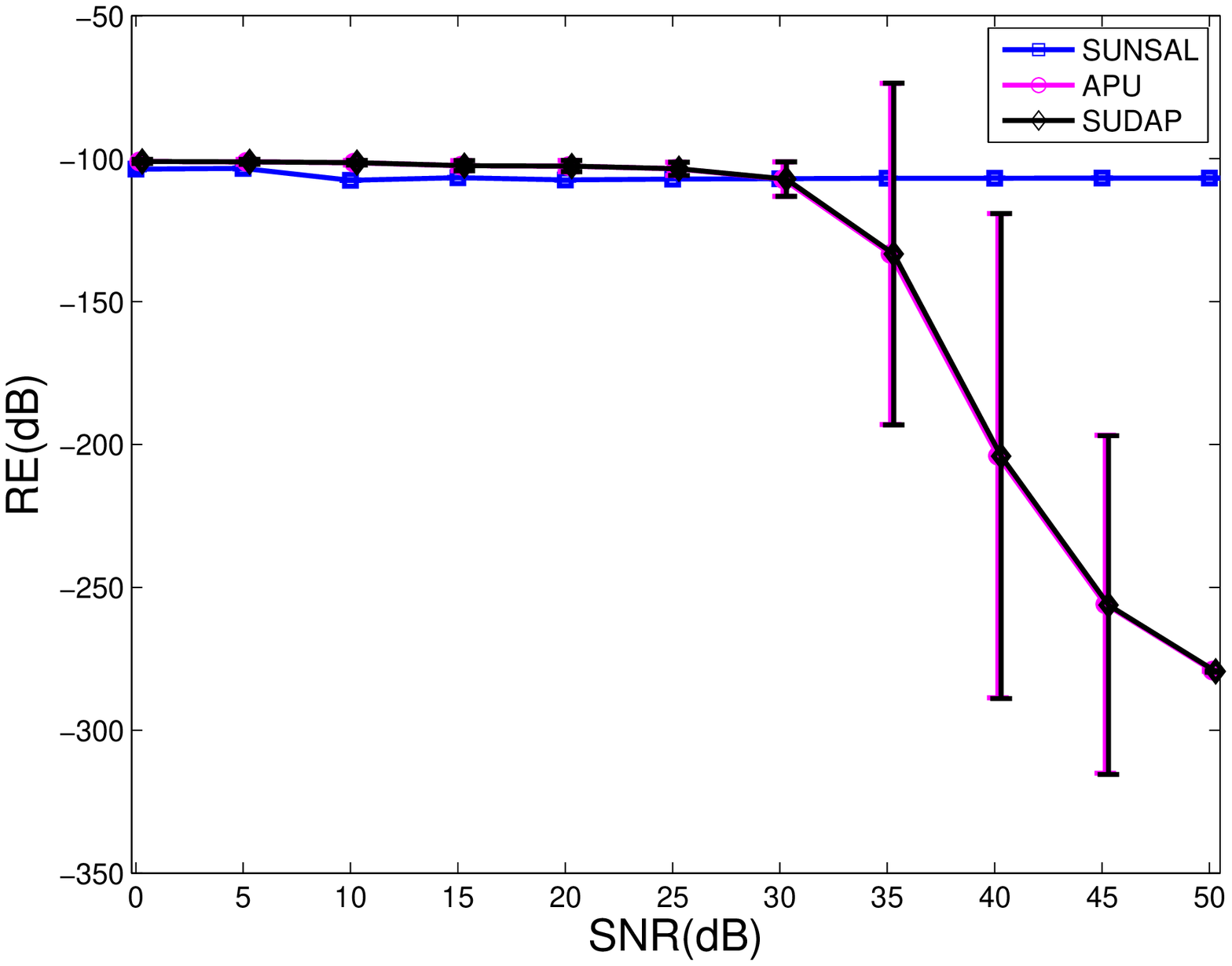}
\includegraphics[width=0.49\columnwidth]{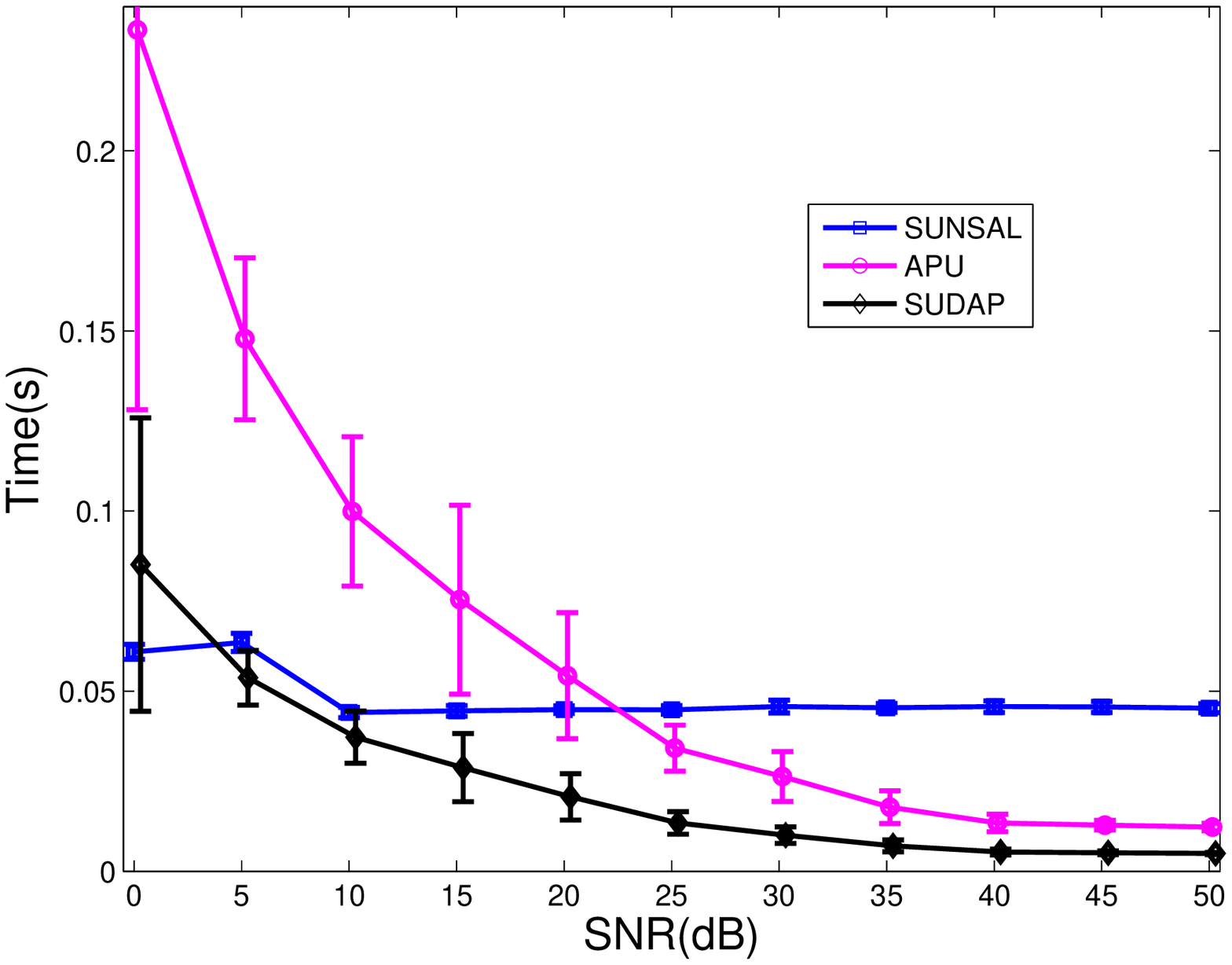}
\caption{RE (left) and time (right) \emph{vs.} SNR for SUNSAL, APU and SUDAP (RE$_t<-100$dB).}
\label{fig:RE_SNR}
\end{figure}

In this experiment, the SNR of the HS image varies from $0$dB to $50$dB
while the other parameters are the same as in Section \ref{subsubsec:perf_time}.
The stopping rule is the one of Section \ref{subsubsec:time_end}.
The results are displayed in Fig. \ref{fig:RE_SNR} and indicate that SUNSAL is more efficient than 
APU and SUDAP (i.e., uses less time) for low SNR scenarios. More specifically, to
achieve RE$_t<-100$dB, SUNSAL provides more efficient unmixing when the SNR is 
lower than $5$dB while SUDAP is faster than SUNSAL when the SNR is higher than $5$dB.

\subsection{Real Data}
\label{subsec:real}
This section compares the performance of the proposed SUDAP algorithm 
with that of SUNSAL and APU using two real datasets associated with two
different applications, i.e., spectroscopy and hyperspectral imaging.


\subsubsection{EELS Dataset}
\label{subsubsec:EELS}
\begin{figure}[h!]
\centering
\begin{minipage}[c]{0.39\columnwidth}
  \includegraphics[width=\columnwidth]{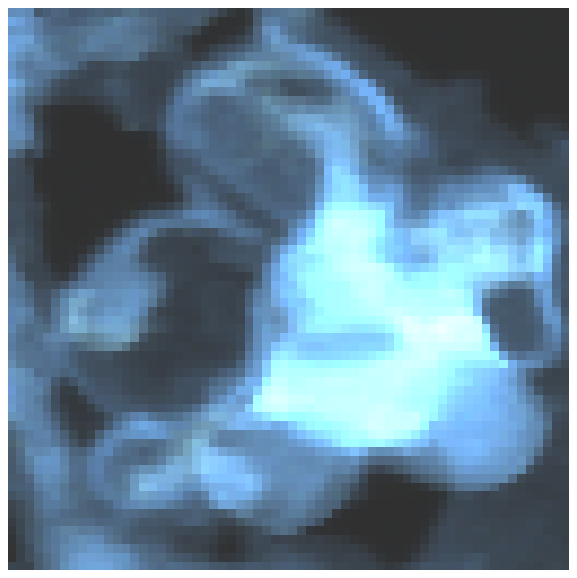}
\end{minipage}
\begin{minipage}[c]{0.59\columnwidth}
  \includegraphics[width=\columnwidth]{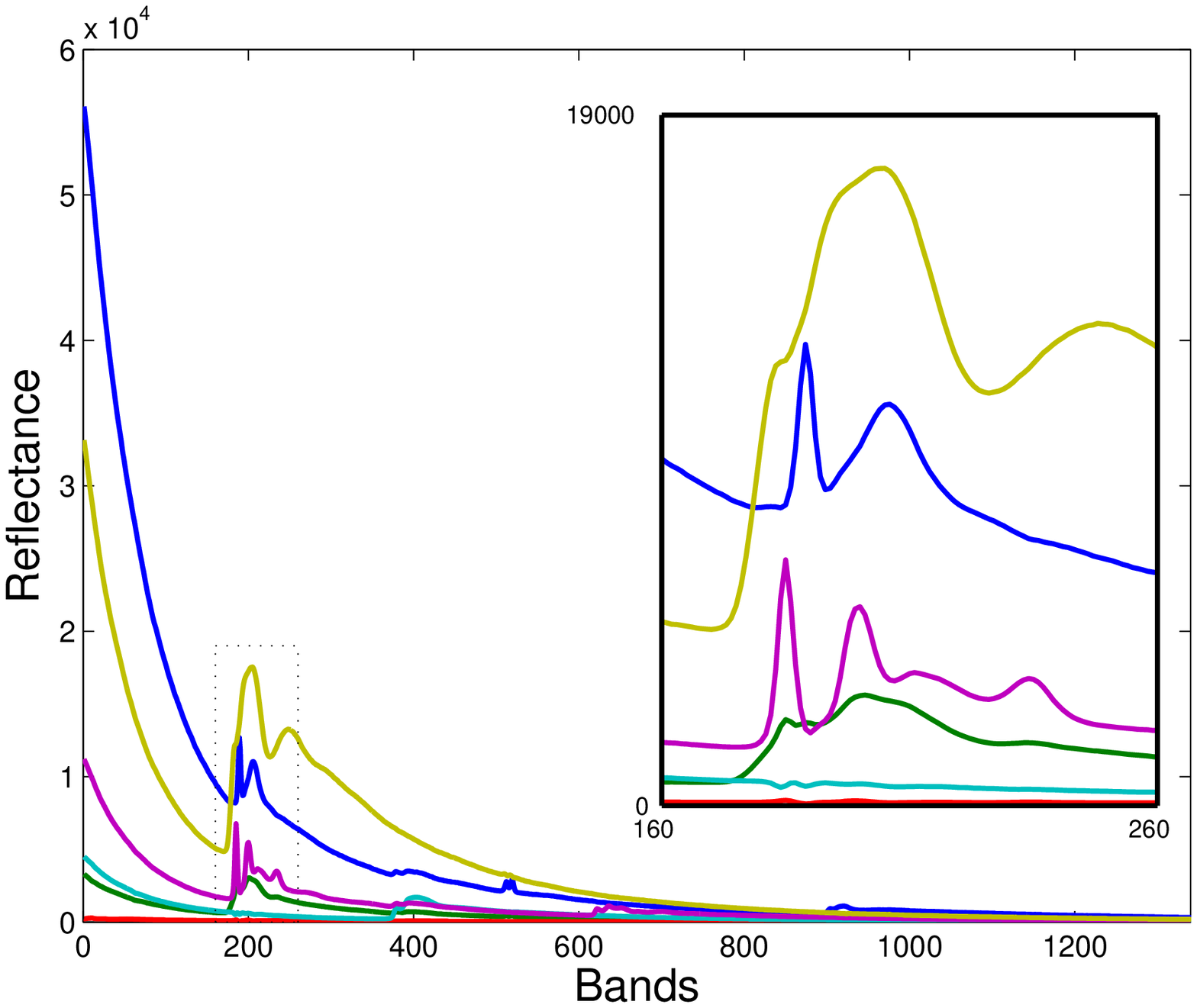}
\end{minipage}
\caption{EELS dataset: HS image (left) and extracted endmember signatures (right).}
\label{fig:HS_EELS}
\end{figure}

\begin{figure*}[t!]
\centering
    \subfigure{
    \includegraphics[width=0.15\textwidth]{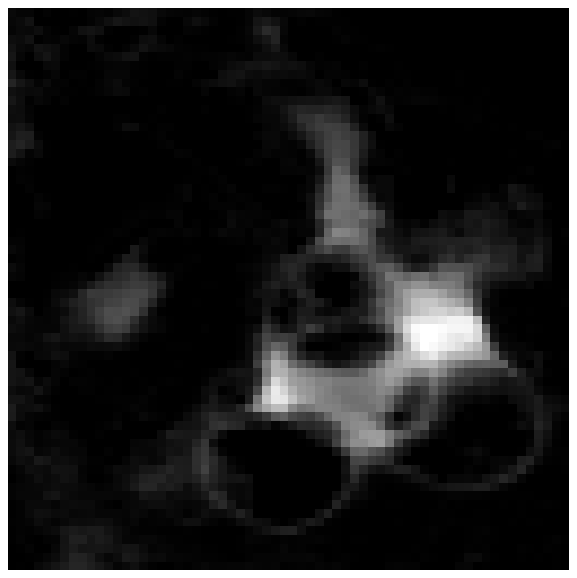}}
    \subfigure{
    \includegraphics[width=0.15\textwidth]{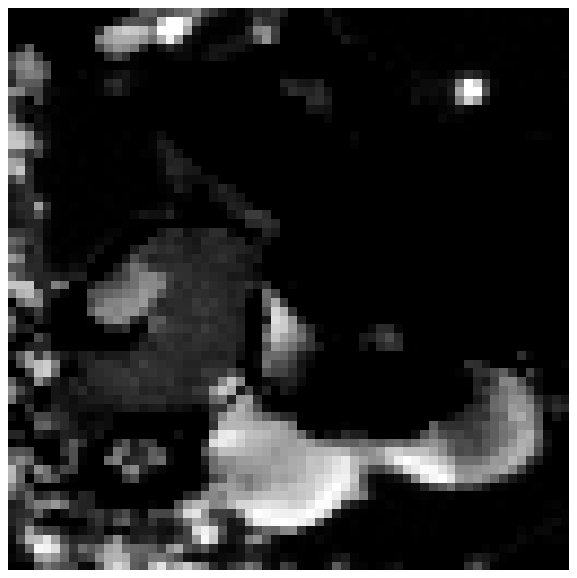}}
    \subfigure{
    \includegraphics[width=0.15\textwidth]{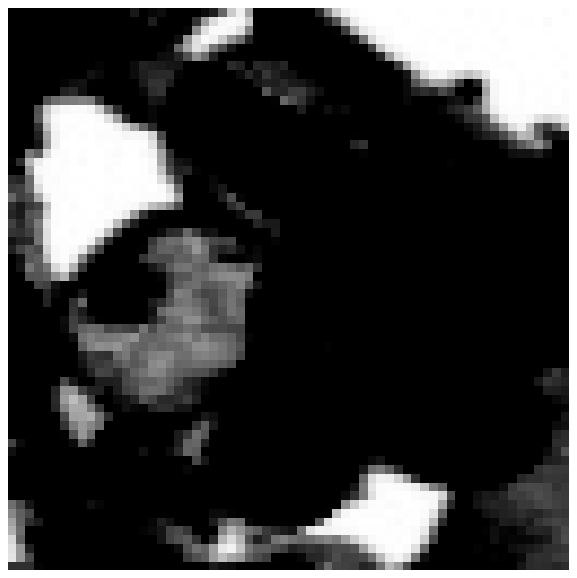}}
    \subfigure{
    \includegraphics[width=0.15\textwidth]{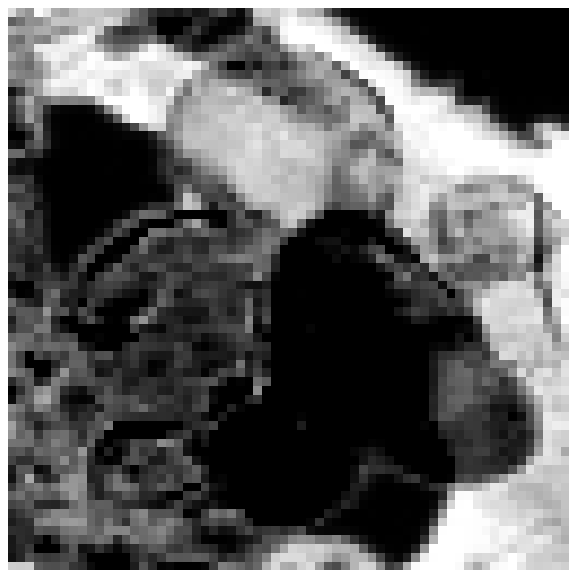}}
    \subfigure{
    \includegraphics[width=0.15\textwidth]{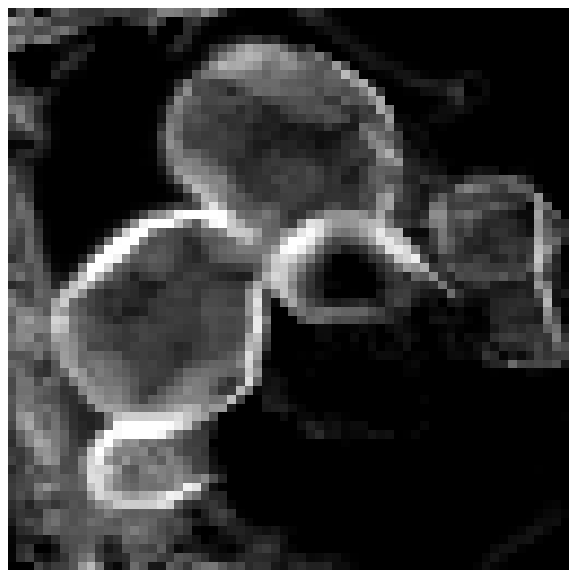}}
    \subfigure{
    \includegraphics[width=0.15\textwidth]{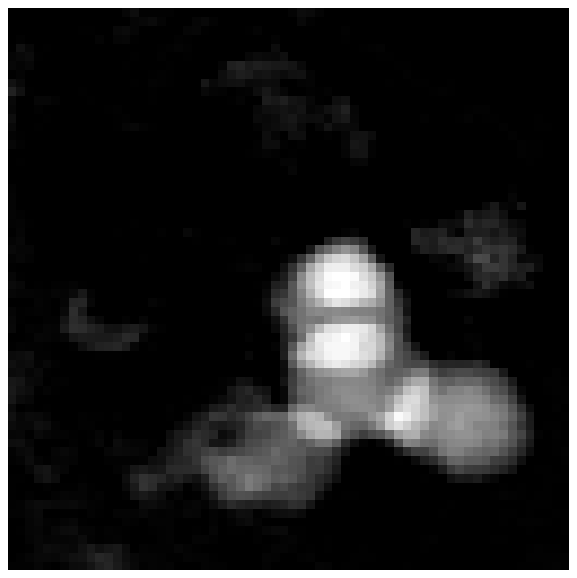}}\\
    \subfigure{
    \includegraphics[width=0.15\textwidth]{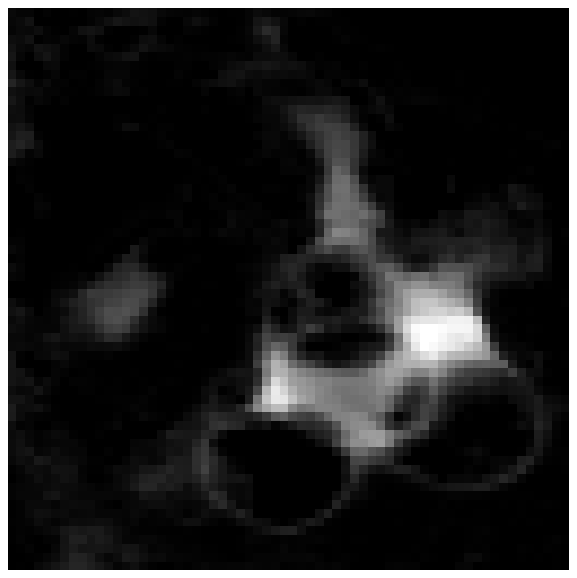}}
    \subfigure{
    \includegraphics[width=0.15\textwidth]{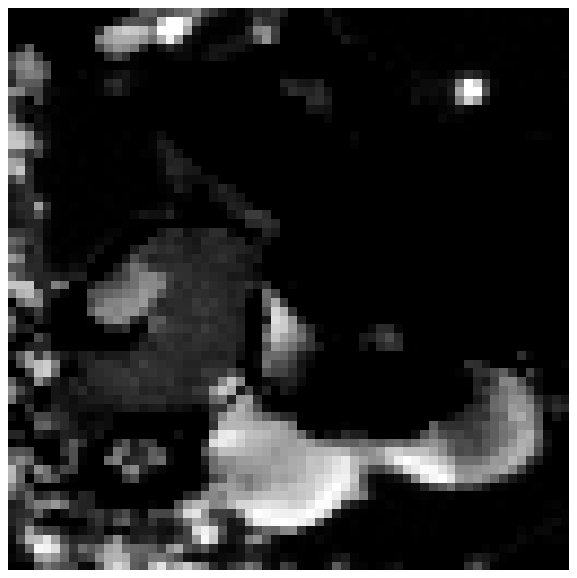}}
    \subfigure{
    \includegraphics[width=0.15\textwidth]{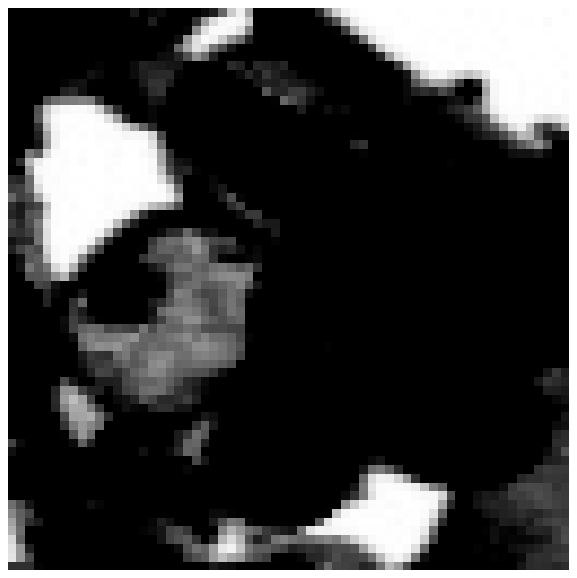}}
    \subfigure{
    \includegraphics[width=0.15\textwidth]{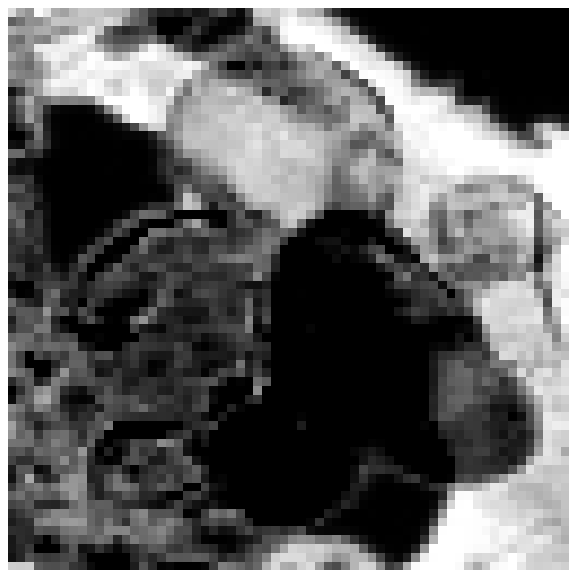}}
    \subfigure{
    \includegraphics[width=0.15\textwidth]{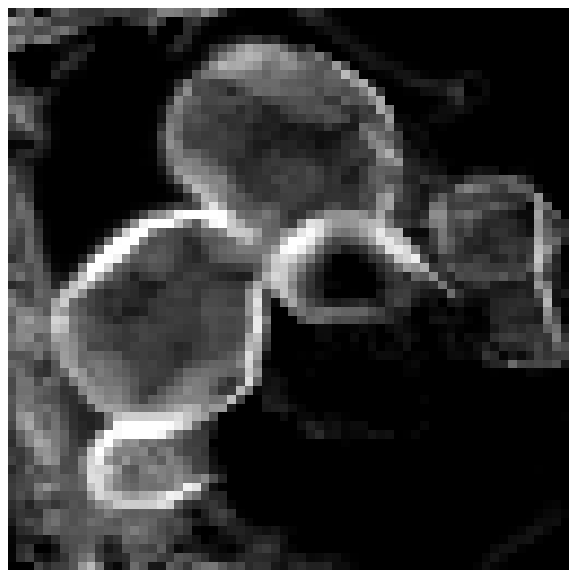}}
    \subfigure{
    \includegraphics[width=0.15\textwidth]{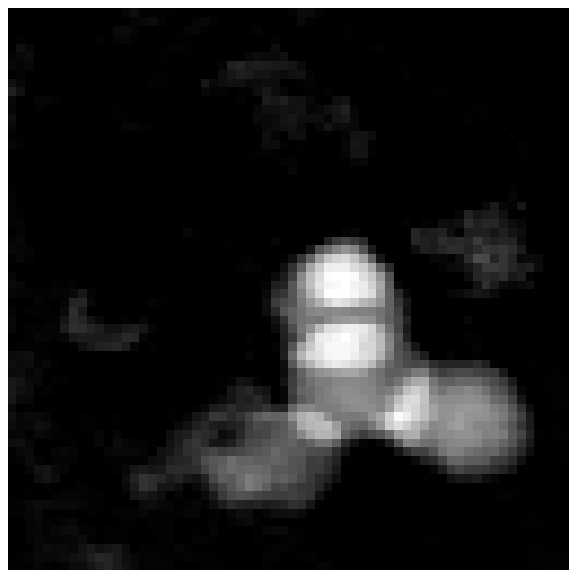}}\\
    \subfigure{
    \includegraphics[width=0.15\textwidth]{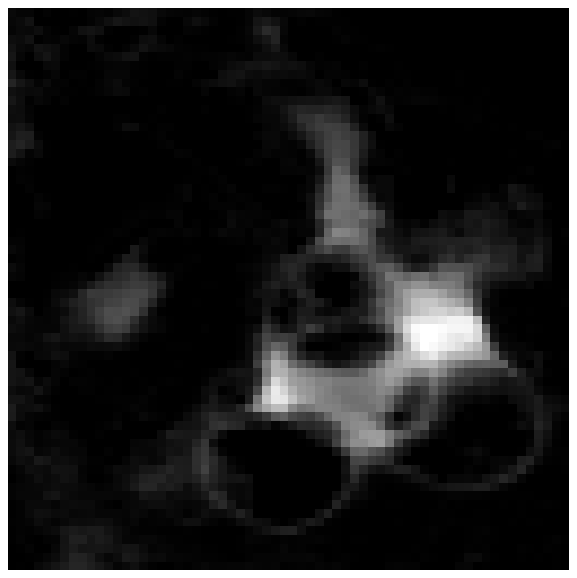}}
    \subfigure{
    \includegraphics[width=0.15\textwidth]{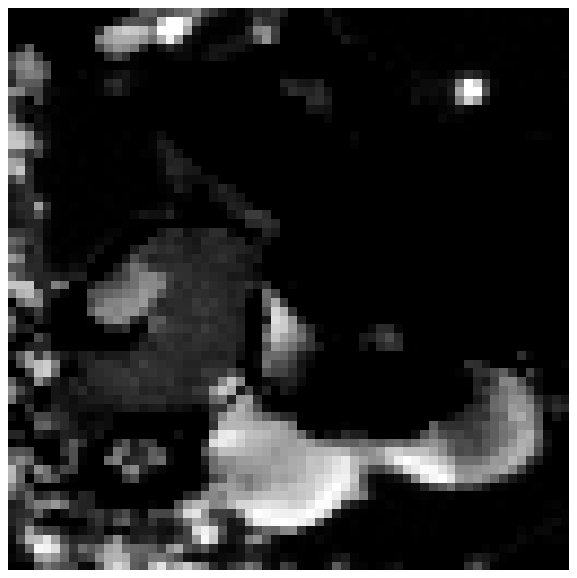}}
    \subfigure{
    \includegraphics[width=0.15\textwidth]{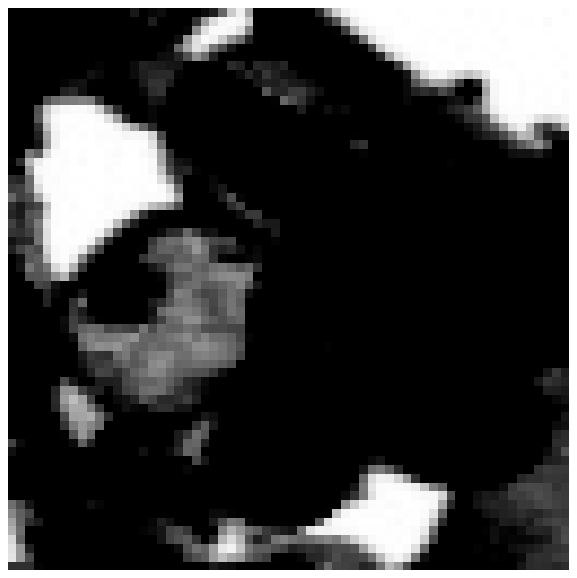}}
    \subfigure{
    \includegraphics[width=0.15\textwidth]{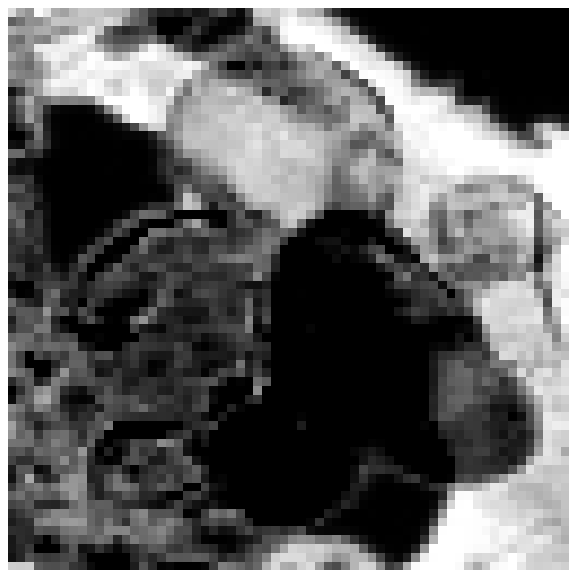}}
    \subfigure{
    \includegraphics[width=0.15\textwidth]{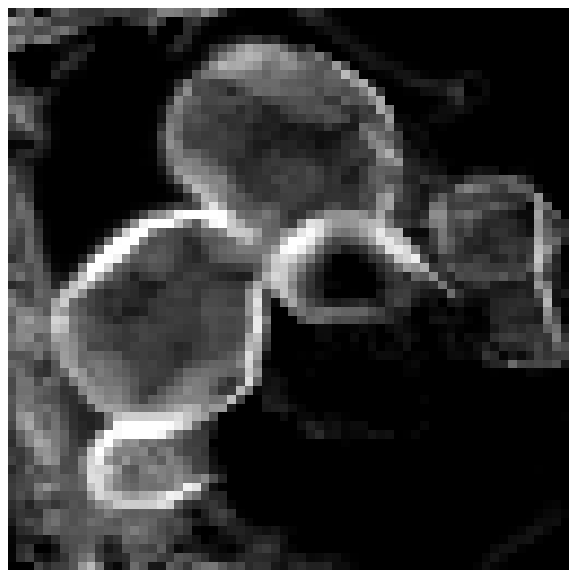}}
    \subfigure{
    \includegraphics[width=0.15\textwidth]{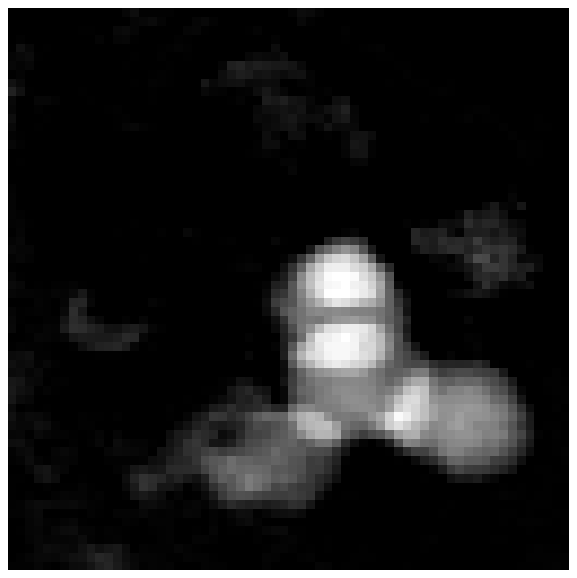}}
        \caption{EELS dataset: abundance maps estimated by SUNSAL (top), APU (middle) and SUDAP (bottom).}
\label{fig:Real_Abu_EELS}
\end{figure*}

In this experiment, a spectral image acquired by electron energy-loss spectroscopy (EELS) is considered.
The analyzed dataset is a $64\times64$ pixel spectrum-image
acquired in $n_{\lambda}=1340$ energy channels over a region composed of several nanocages
in a boron-nitride nanotubes (BNNT) sample \cite{Dobigeon2012ultra}.
A false color image of the EELS data (with an arbitrary selection of three channels as
RGB bands) is displayed in Fig. \ref{fig:HS_EELS} (left). Following \cite{Dobigeon2012ultra}, the number
of endmembers has been set to $m=6$. The endmember signatures have been
extracted from the dataset using VCA \cite{Nascimento2005} and are depicted in Fig. \ref{fig:HS_EELS} (right). The abundance maps estimated by the considered unmixing algorithms are shown in Fig. \ref{fig:Real_Abu_EELS} for a stopping rule defined as RE$_t<100$dB.


\begin{figure}[h!]
\centering
\includegraphics[width=0.49\columnwidth]{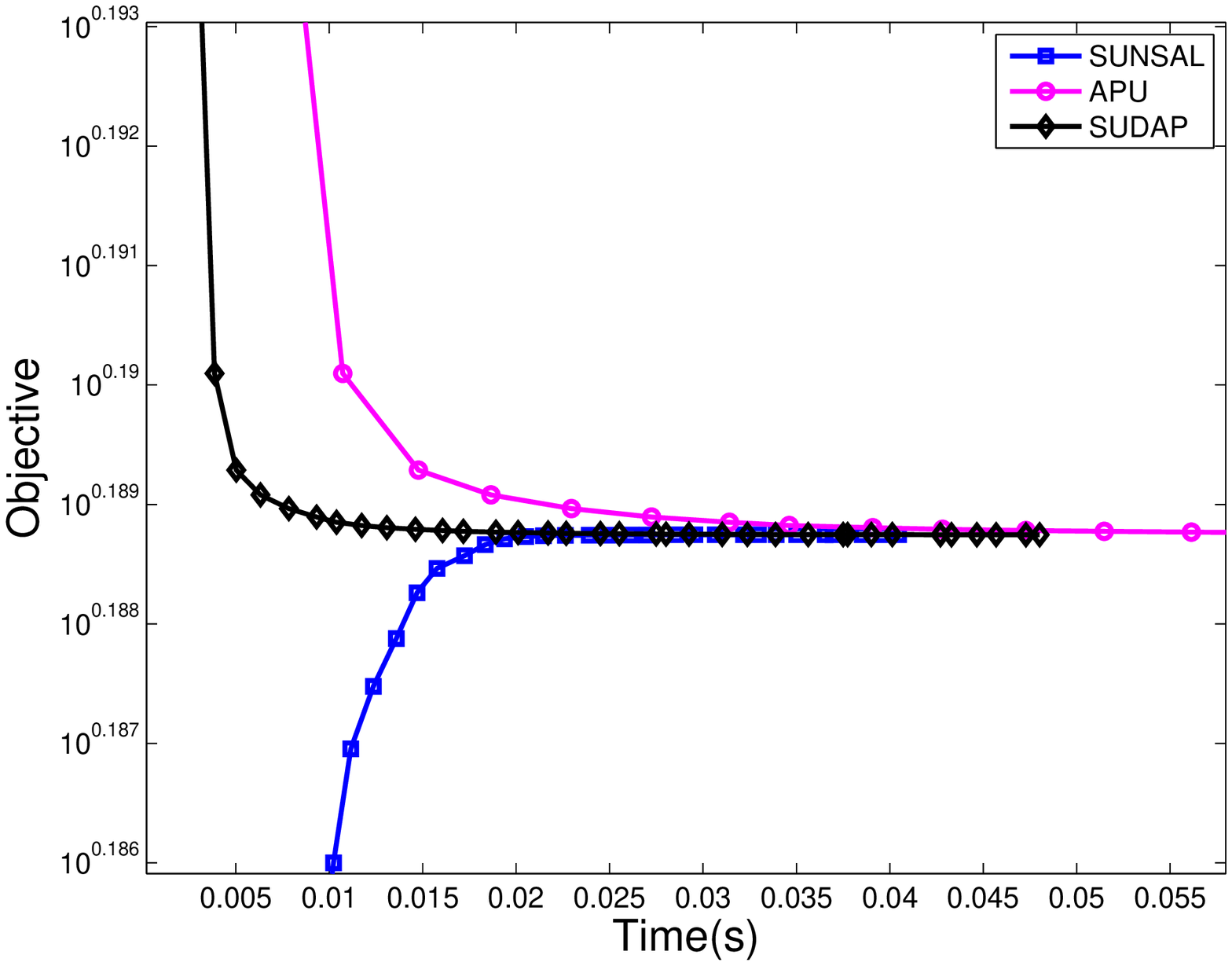}
\includegraphics[width=0.49\columnwidth]{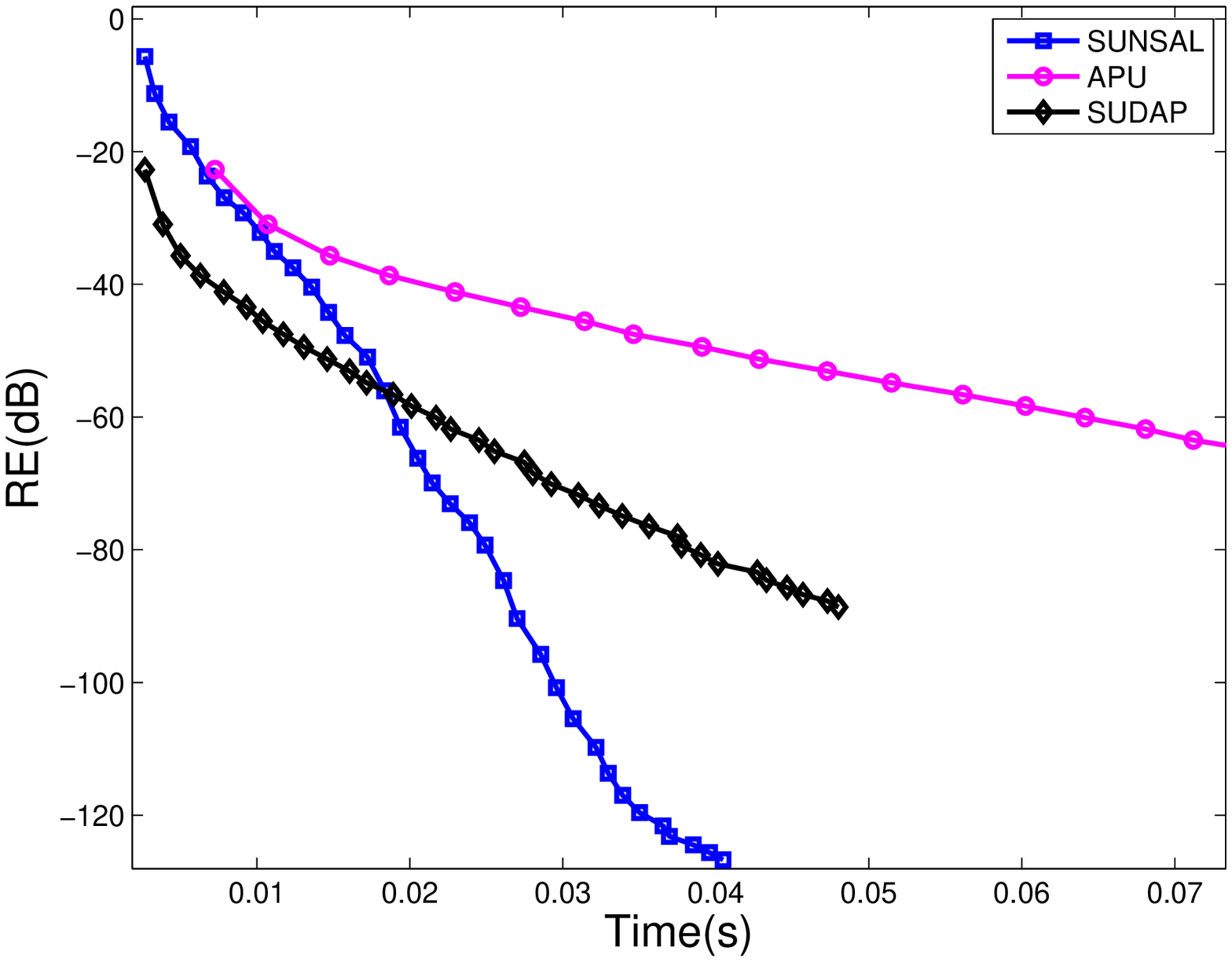}
\caption{Objective (left) and RE (right) \emph{vs.} time for SUNSAL, APU and SUDAP (EELS data).}
\label{fig:Real_Obj_EELS}
\end{figure}

There is no visual difference between the abundance maps provided by SUNSAL, APU and the proposed SUDAP.
Since there is no available ground-truth for the abundances, the objective criterion $\calJ_t=\|\bfX-\bfE\hat{\bfA}_t\|_F^2$ minimized by the algorithms has been evaluated instead of
NMSE$_t$. The variations of the objective function and the corresponding REs are
displayed in Fig. \ref{fig:Real_Obj_EELS} as a function of the computational time.

\begin{figure}[h!]
\centering
\includegraphics[width=0.49\columnwidth]{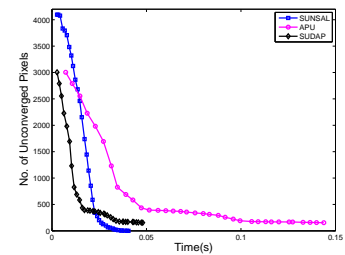}
\caption{Number of pixels that do not satisfy the stopping rule \emph{vs.} time for SUNSAL, APU and SUDAP (EELS data).}
\label{fig:Real_NumUncvg_EELS}
\end{figure}

Both figures show that the proposed SUDAP performs faster than APU and SUNSAL as long as the stopping rule has been fixed
as RE$_t<-60$dB. For lower RE$_t$, SUDAP becomes less efficient than SUNSAL. To explore the convergence more explicitly, 
the number of spectral vectors that do not satisfy the convergence criterion, i.e., for which RE$>-100$dB, has been determined
and is depicted in Fig. \ref{fig:Real_NumUncvg_EELS}.
It is clear that most of the spectral vectors (around $3600$ out of $4096$ pixels) converged
quickly, e.g., in less than $0.02$ seconds. The remaining measurements (around $500$ pixels) require longer time
to converge, which leads to the slow convergence as observed in Fig. \ref{fig:Real_Obj_EELS}.
The slow convergence of the projection methods for these pixels may result from an inappropriate observational model due to, e.g., 
endmember variability \cite{Zare2014} or nonlinearity effects \cite{Dobigeon2014}. On the contrary, SUNSAL is more robust to these 
discrepancies and converges faster for these pixels. This corresponds to the results shown in Fig. \ref{fig:Real_Obj_EELS}.

\subsubsection{Cuprite Dataset}
\label{subsubsec:cuprite}
\begin{figure}[h!]
\centering
\begin{minipage}[c]{0.39\columnwidth}
  \includegraphics[width=\columnwidth]{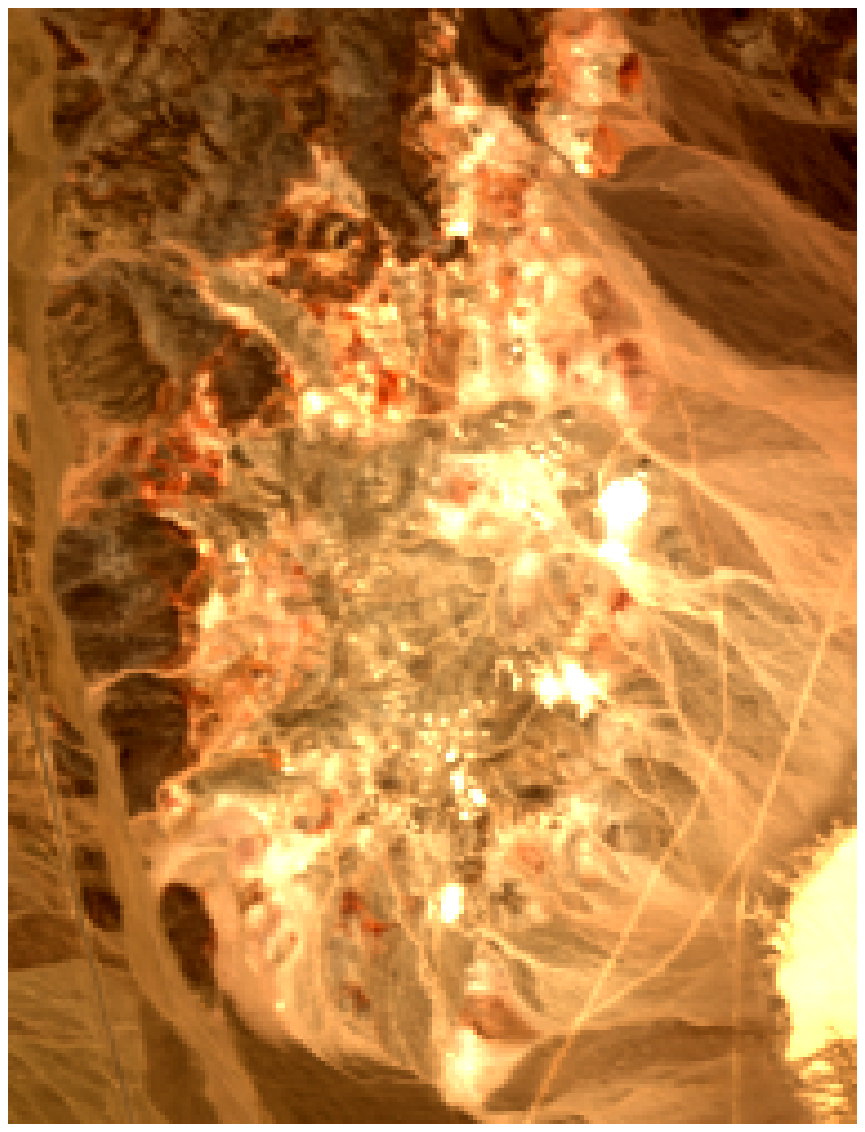}
\end{minipage}
\begin{minipage}[c]{0.59\columnwidth}
  \includegraphics[width=\columnwidth]{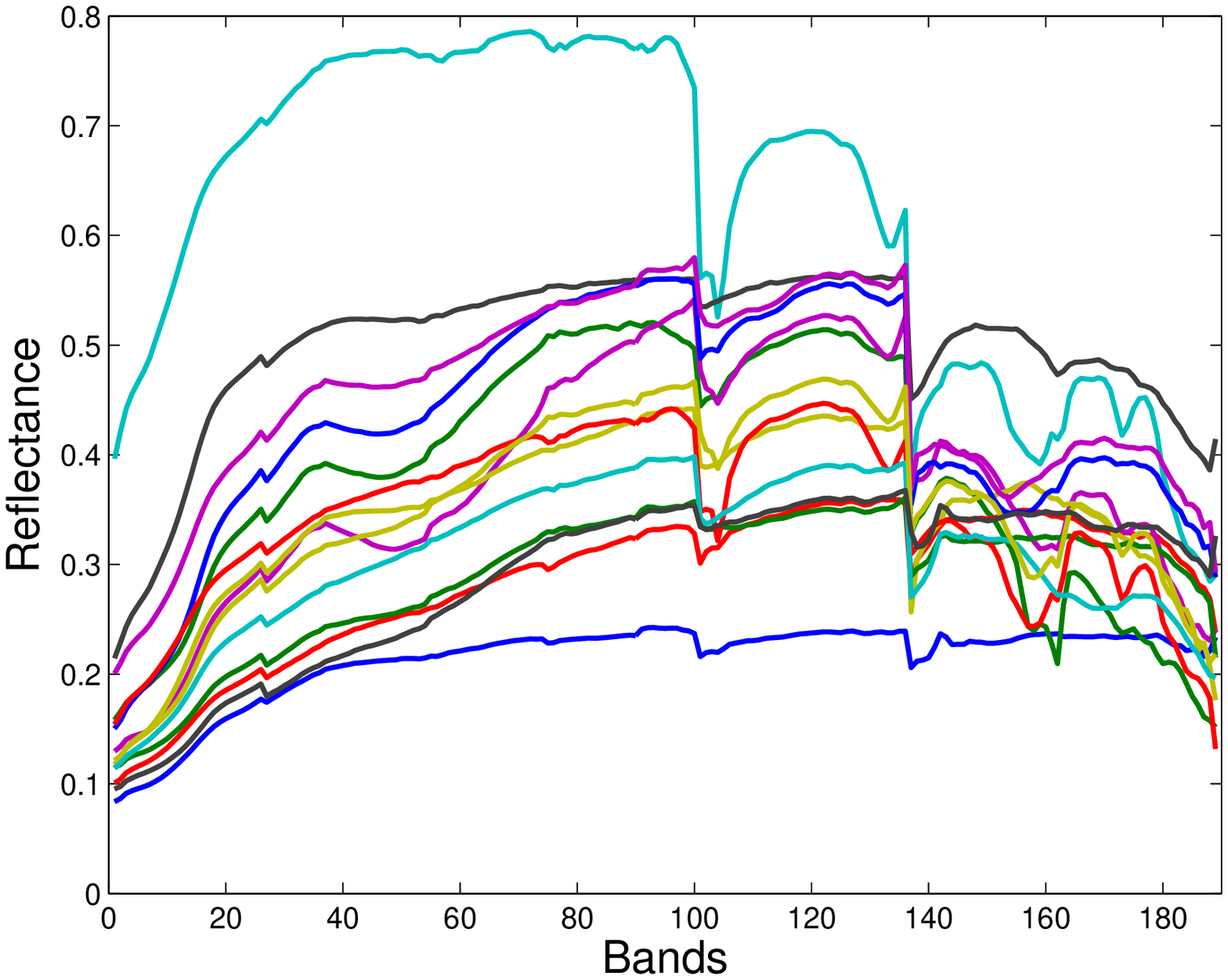}
\end{minipage}
\caption{Cuprite dataset: HS image (left) and extracted endmember signatures (right).}
\label{fig:HS_Cuprite}
\end{figure}

\begin{figure*}[t!]
\centering
    \subfigure{
    \includegraphics[width=0.18\textwidth]{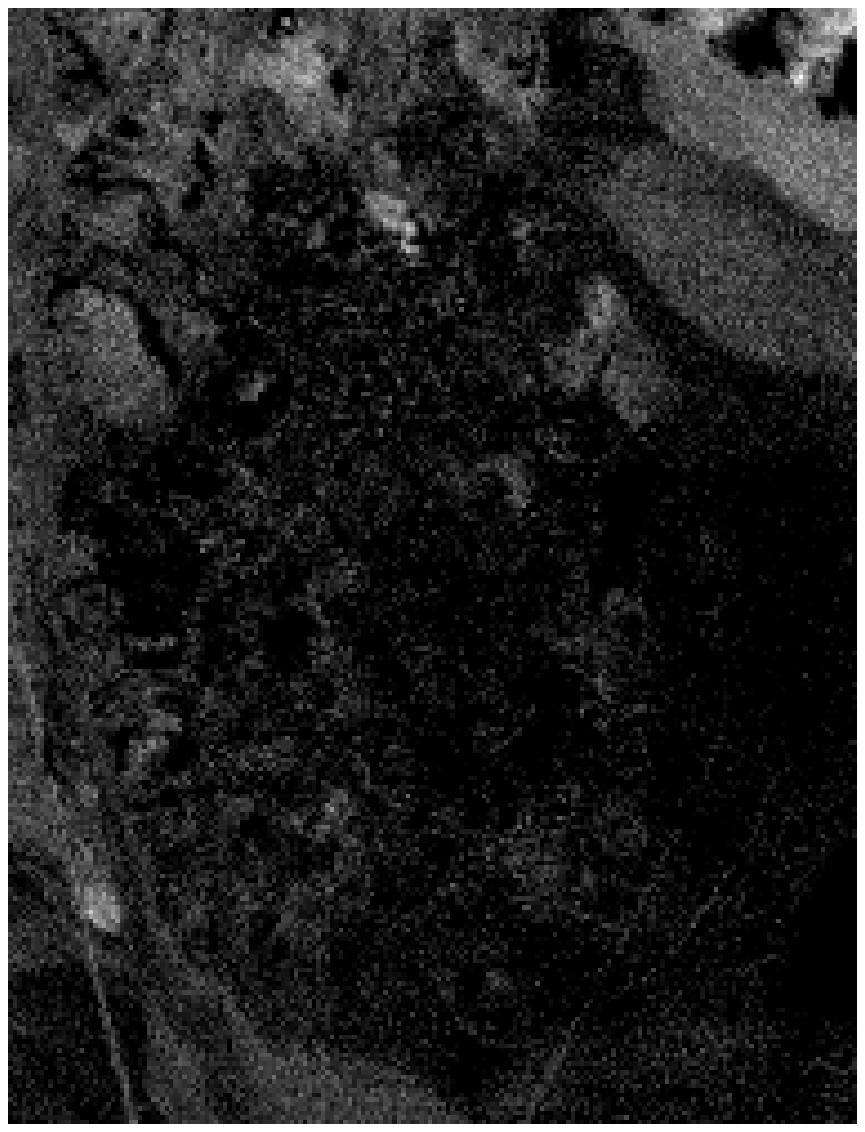}}
    \subfigure{
    \includegraphics[width=0.18\textwidth]{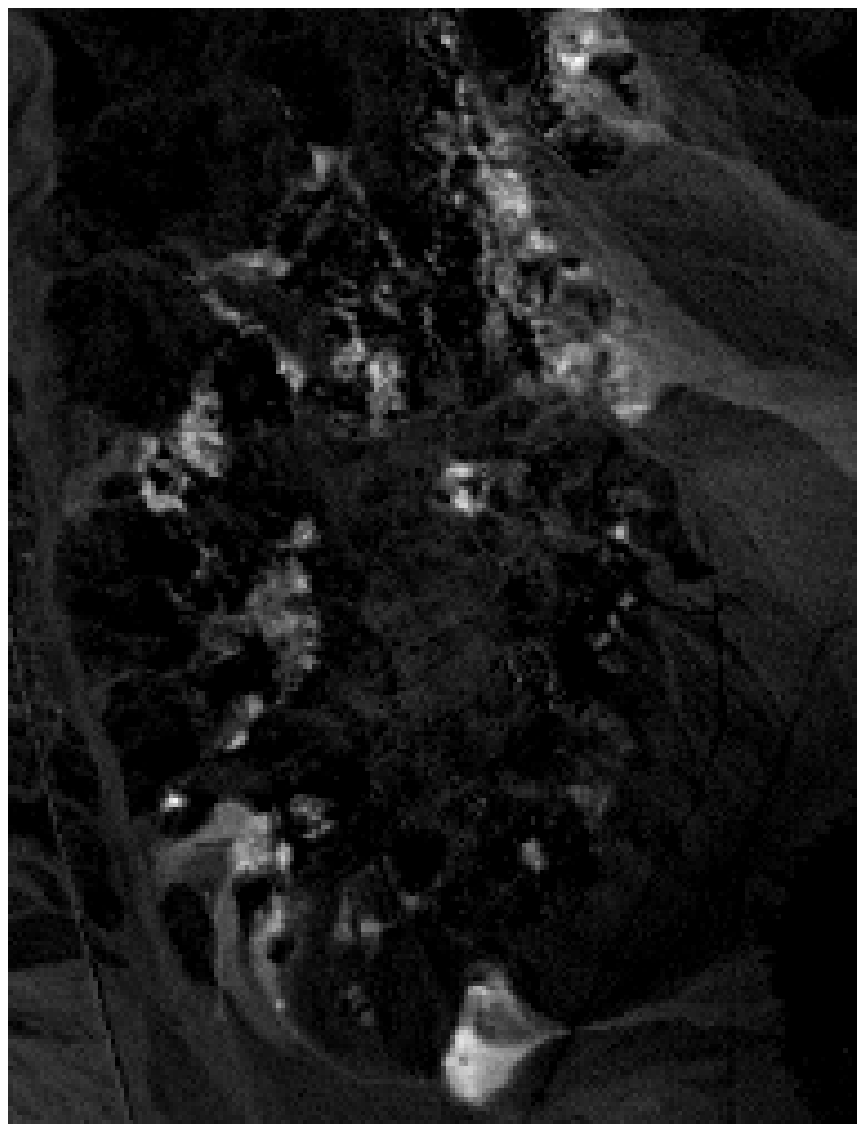}}
    \subfigure{
    \includegraphics[width=0.18\textwidth]{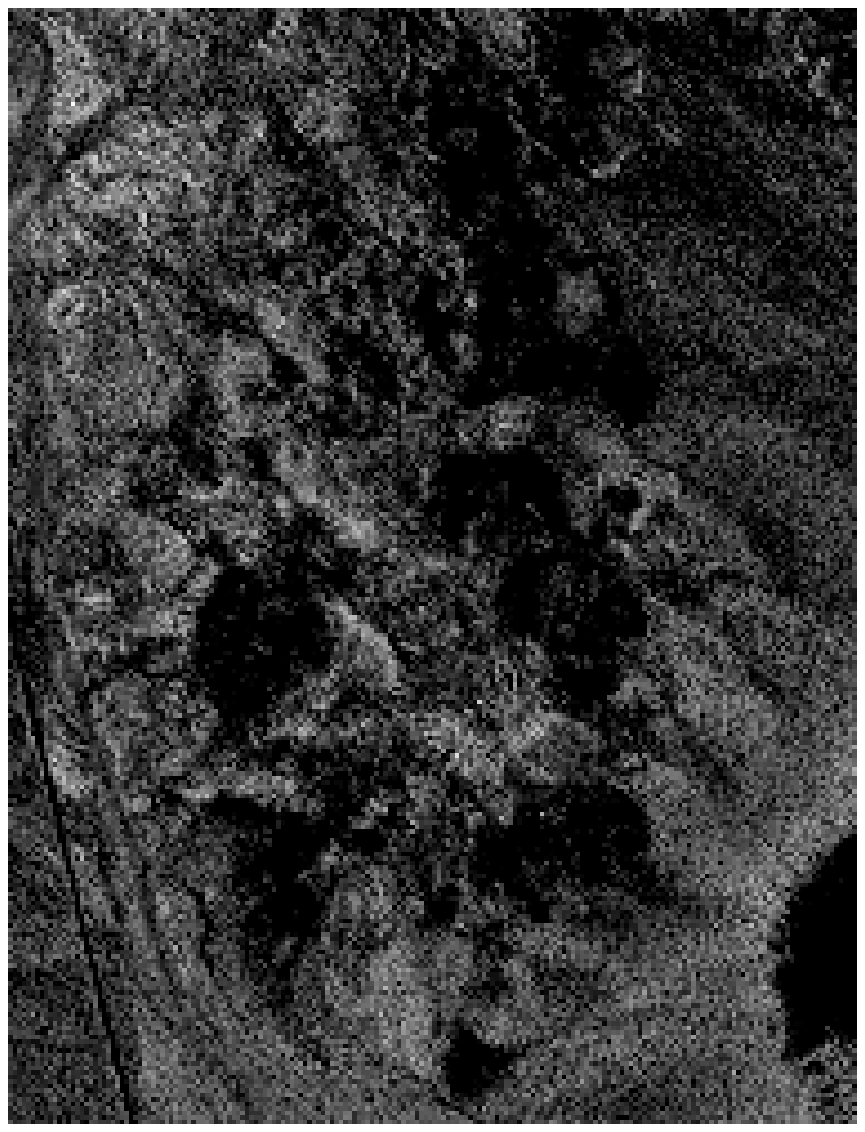}}
    \subfigure{
    \includegraphics[width=0.18\textwidth]{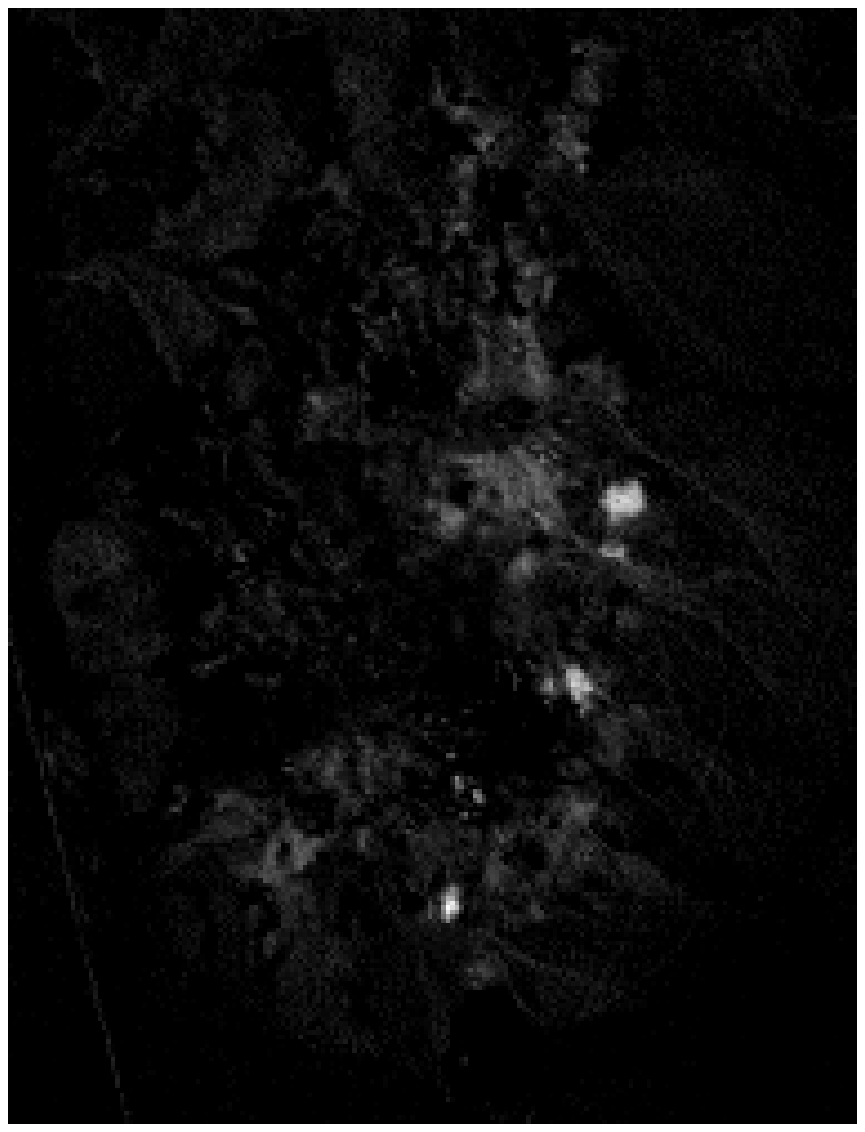}}
    \subfigure{
    \includegraphics[width=0.18\textwidth]{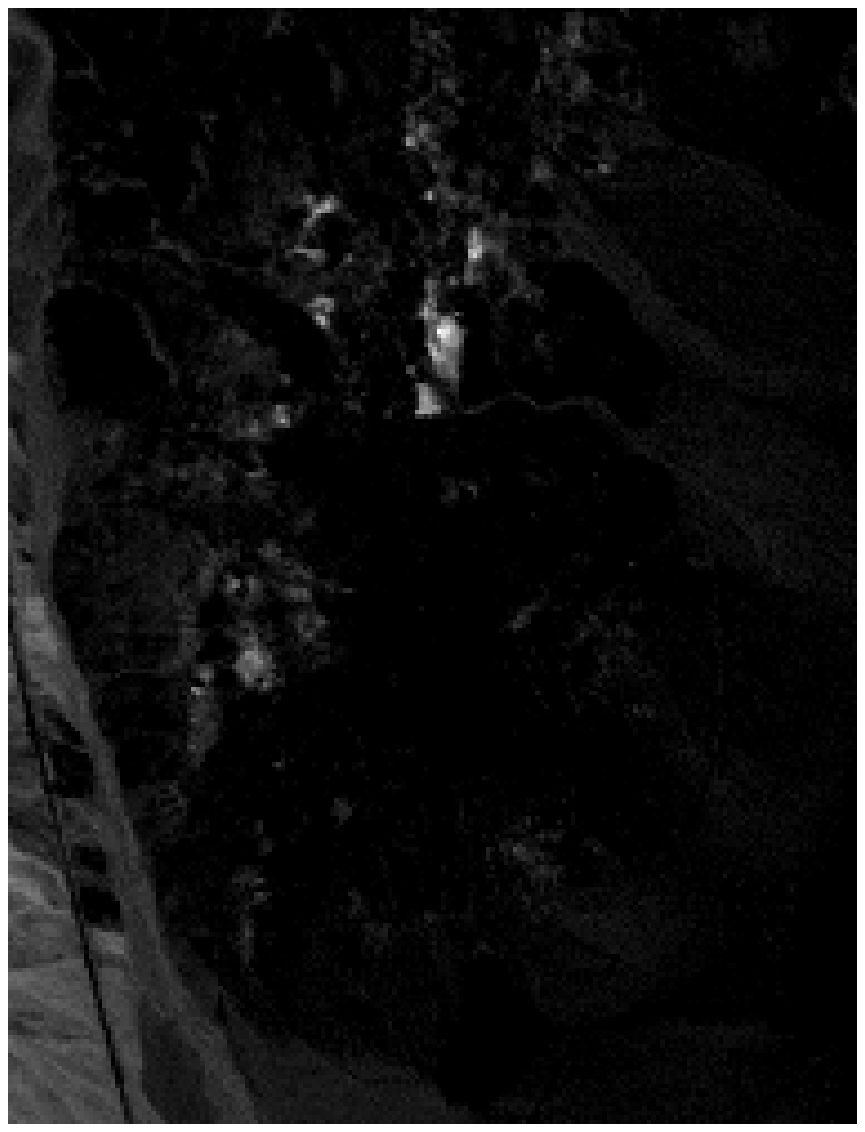}}\\
    \subfigure{
    \includegraphics[width=0.18\textwidth]{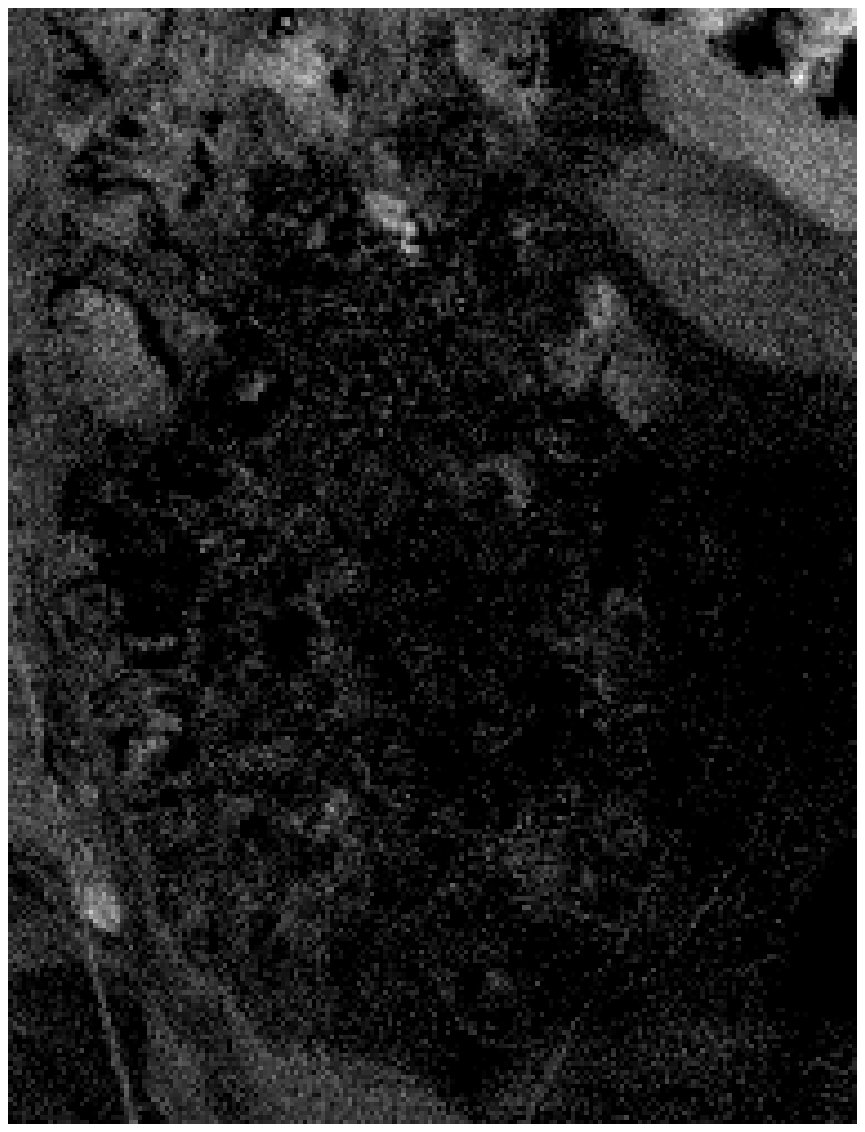}}
    \subfigure{
    \includegraphics[width=0.18\textwidth]{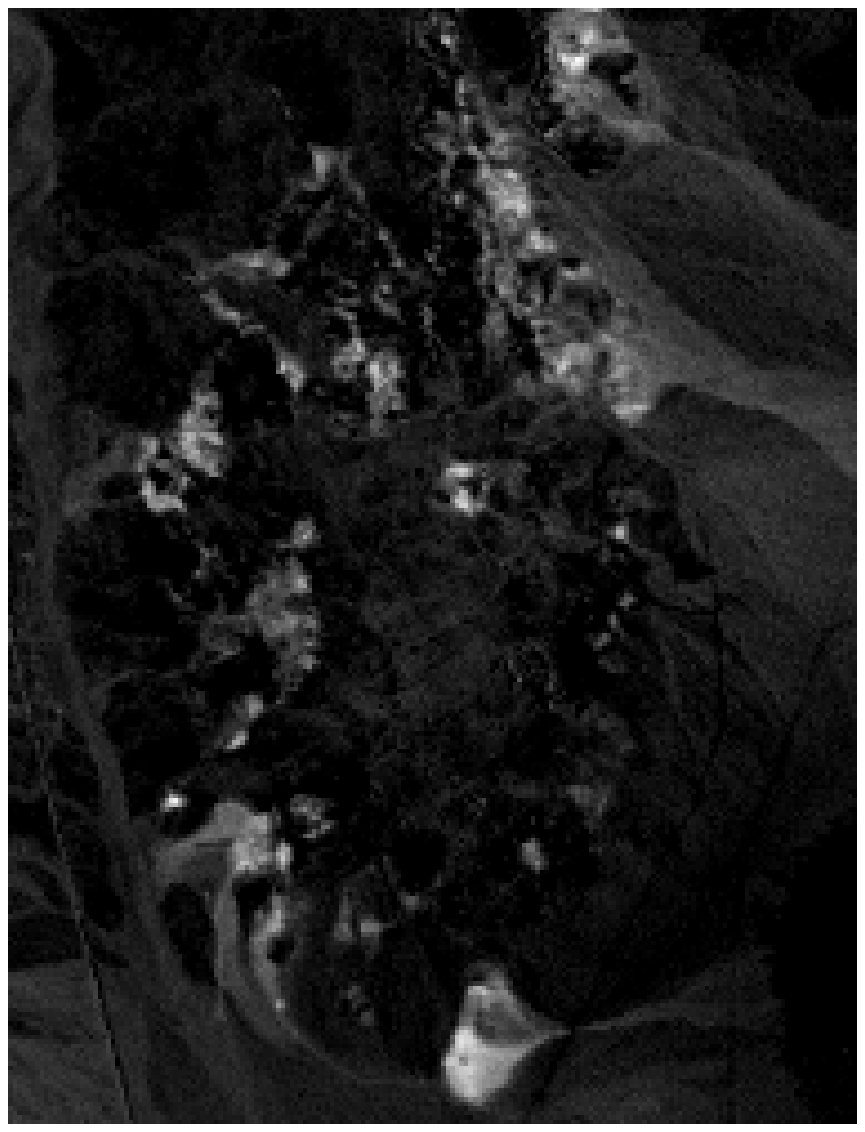}}
    \subfigure{
    \includegraphics[width=0.18\textwidth]{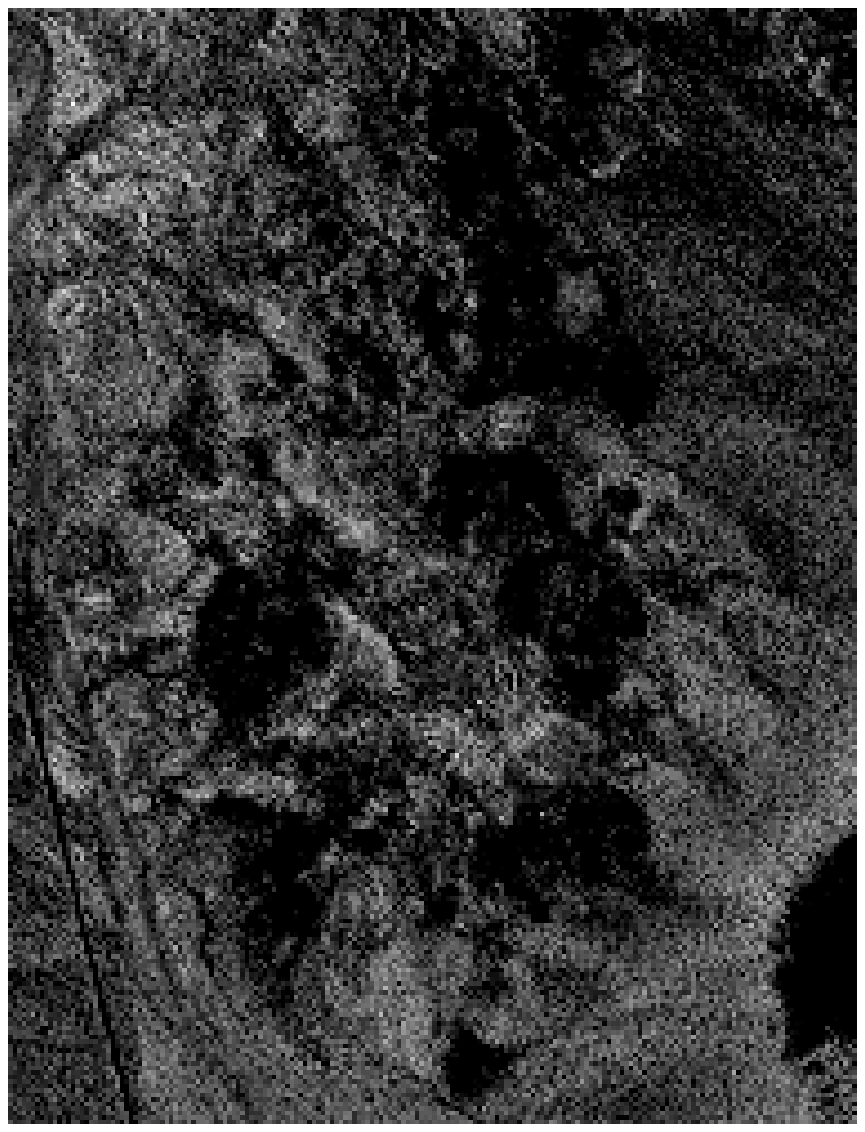}}
    \subfigure{
    \includegraphics[width=0.18\textwidth]{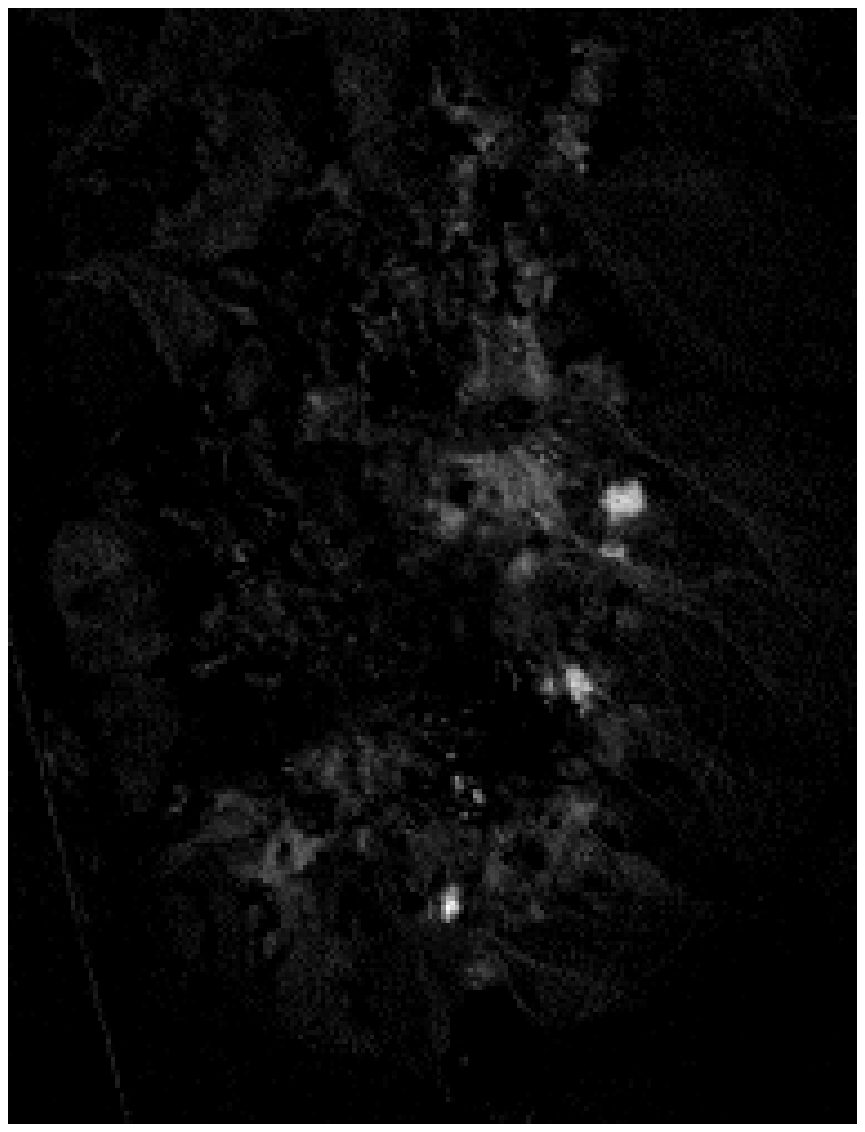}}
    \subfigure{
    \includegraphics[width=0.18\textwidth]{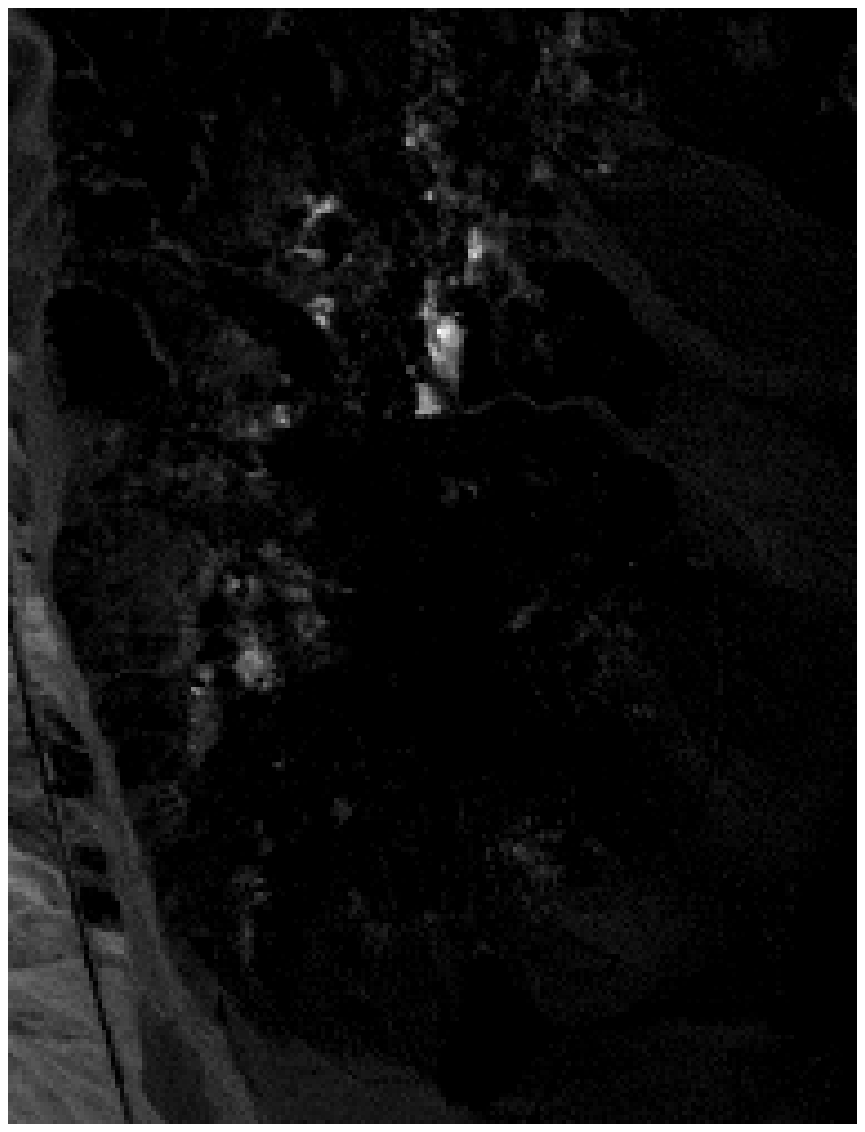}}\\
    \subfigure{
    \includegraphics[width=0.18\textwidth]{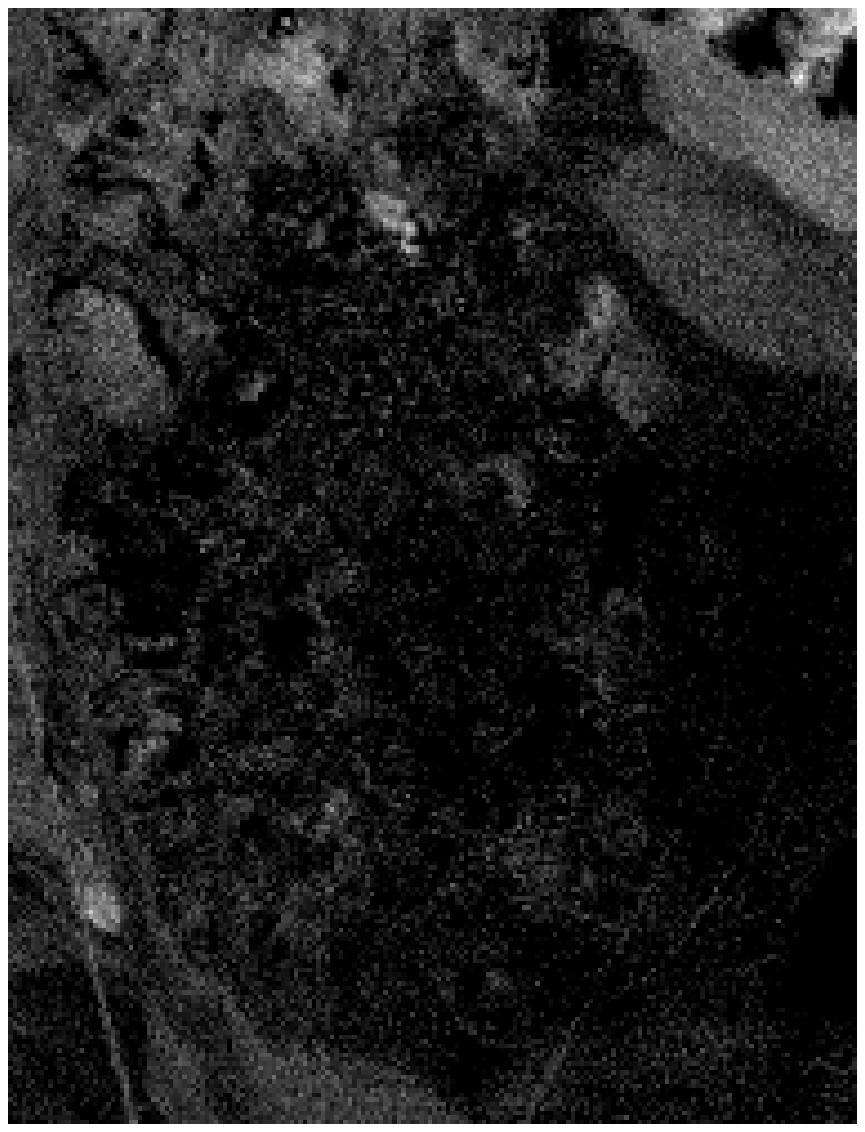}}
    \subfigure{
    \includegraphics[width=0.18\textwidth]{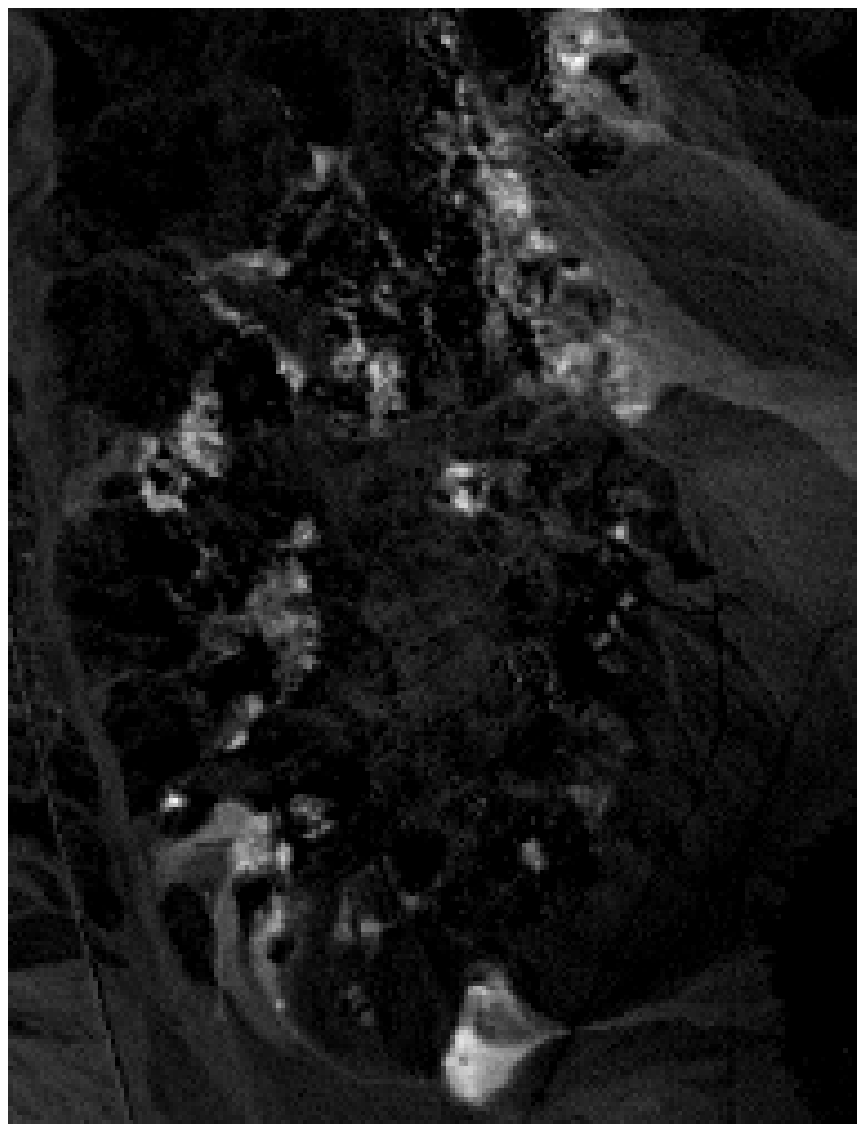}}
    \subfigure{
    \includegraphics[width=0.18\textwidth]{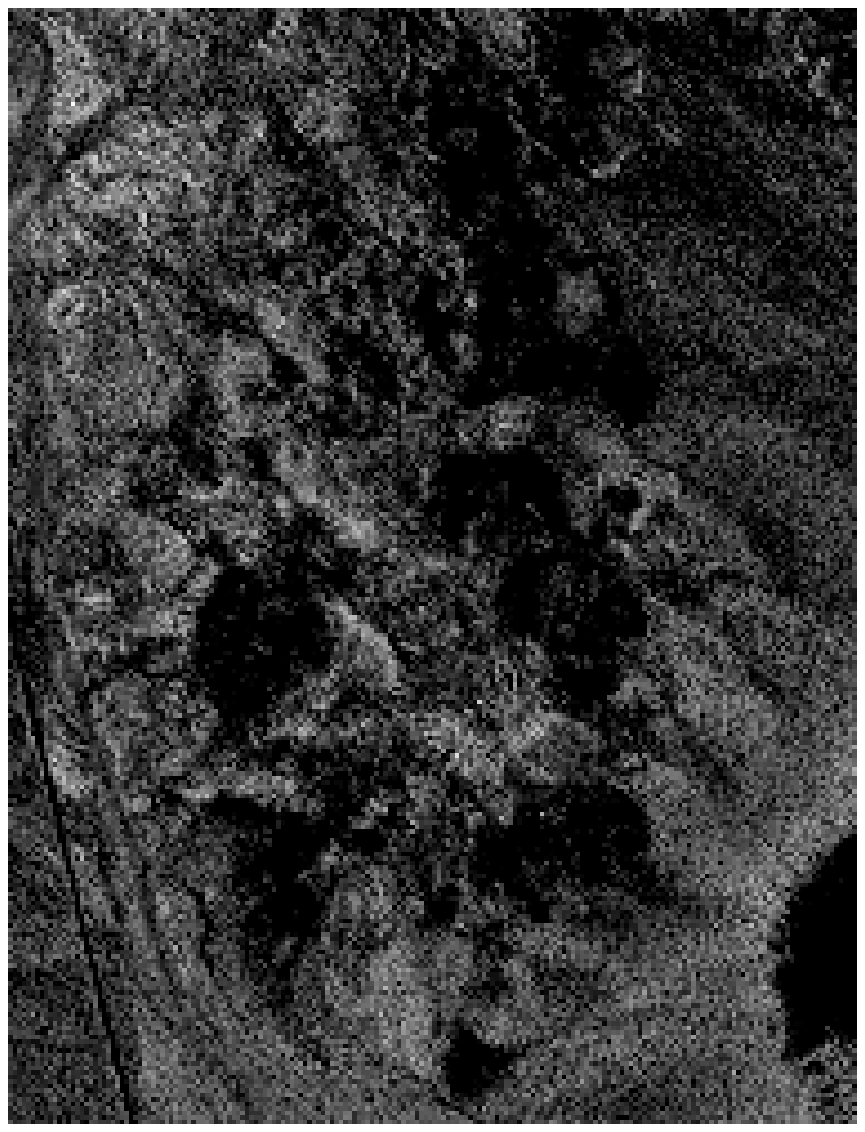}}
    \subfigure{
    \includegraphics[width=0.18\textwidth]{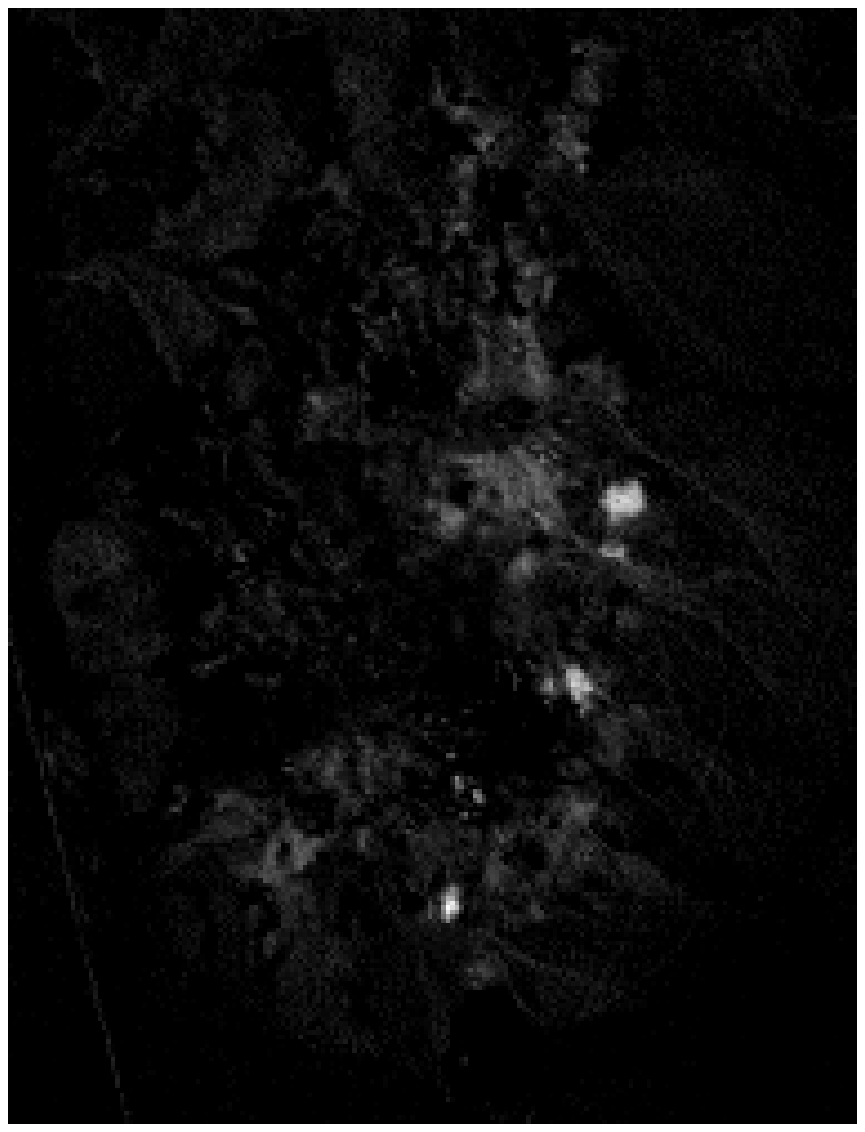}}
      \subfigure{
    \includegraphics[width=0.18\textwidth]{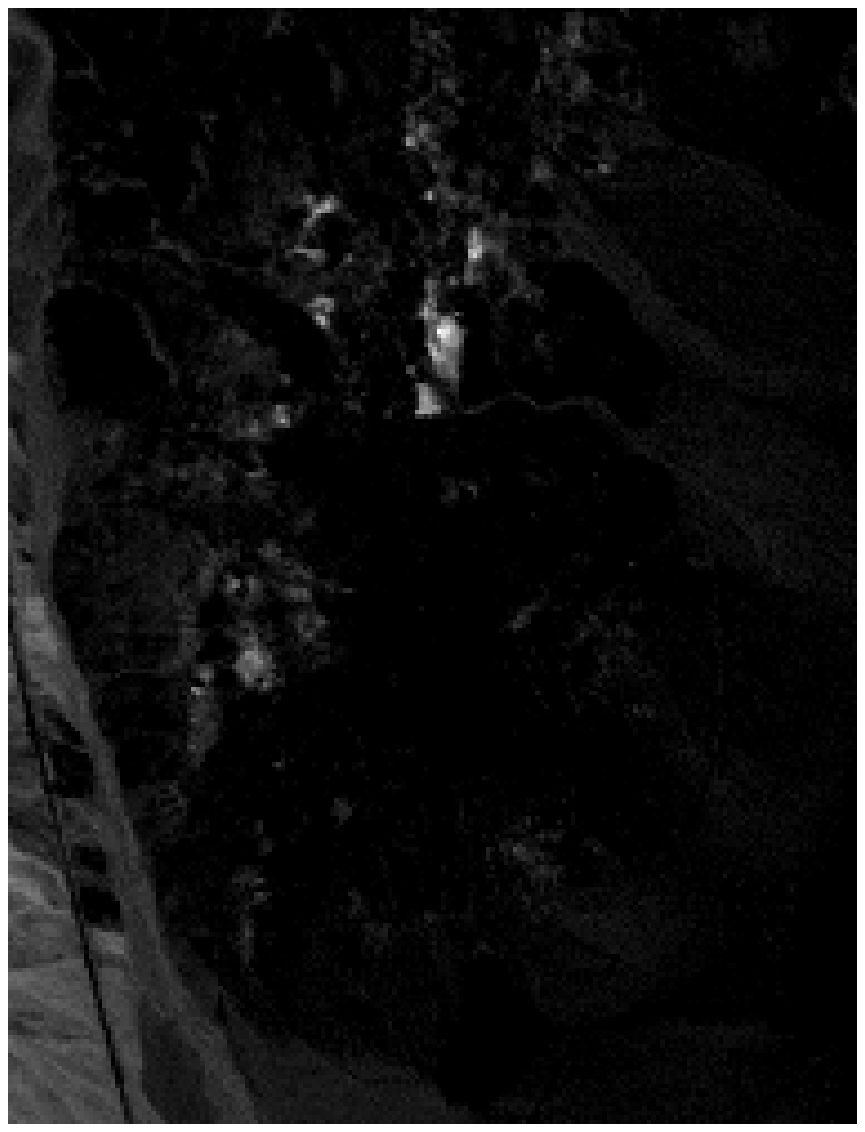}}
    \caption{Cuprite dataset: abundance maps estimated by SUNSAL (top), APU (middle) and SUDAP (bottom).}
\label{fig:Real_Abu_Cuprite}
\end{figure*}

This section investigates the performance of the proposed SUDAP algorithm
when unmixing a real HS image. This image, which has received a lot of interest in
the remote sensing and geoscience literature, was acquired over Cuprite field by the JPL/NASA airborne visible/infrared imaging spectrometer (AVIRIS) \cite{JPL_AVIRIS}.
Cuprite scene is a mining area in southern Nevada composed of several minerals and
some vegetation, located approximately $200$km northwest of Las
Vegas.
The image considered in this experiment consists of $250 \times 190$ pixels of  $n_{\lambda} =189$ spectral bands 
obtained after removing the water vapor absorption bands.
A composite color image of the scene of interest is shown in Fig. \ref{fig:HS_Cuprite} (left).
As in Section \ref{subsubsec:cuprite}, the endmember matrix $\bfE$ has been learnt from
the HS data using VCA. According to \cite{Nascimento2005}, the number
of endmembers has been set to $m=14$. The estimated endmember signatures
are displayed in Fig. \ref{fig:HS_Cuprite} (right) ant the first five corresponding abundance maps recovered by SUNSAL, APU and
SUDAP are shown in Fig. \ref{fig:Real_Abu_Cuprite}. Visually, all three methods provide similar
abundance maps\footnote{Similar results were also observed for abundance maps of the other endmembers.
They are not shown here for brevity and are available in a separate technical report \cite{Wei2015_TechRep_SUDAP}.}. 

\begin{figure}[h!]
\centering
\includegraphics[width=0.49\columnwidth]{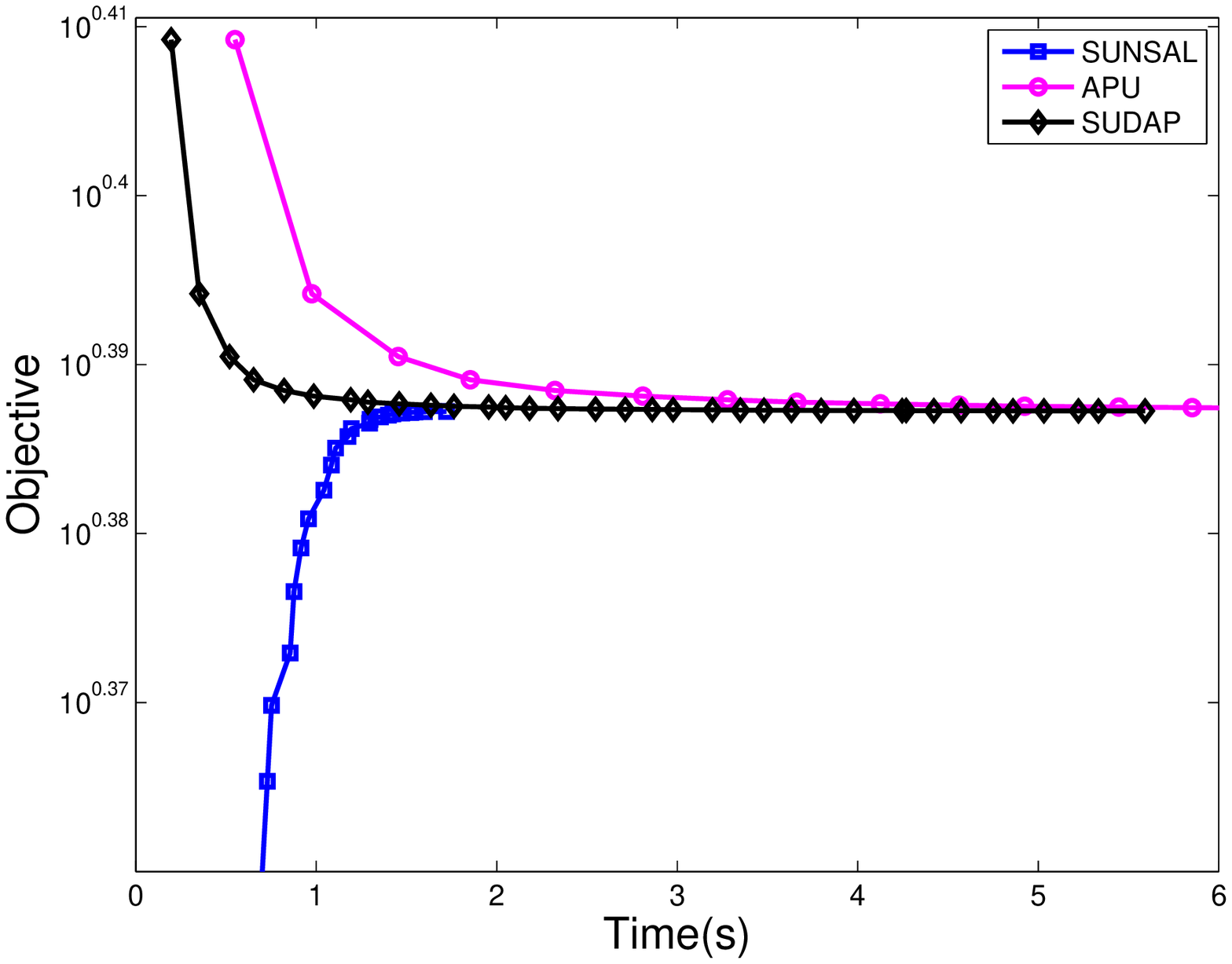}
\includegraphics[width=0.49\columnwidth]{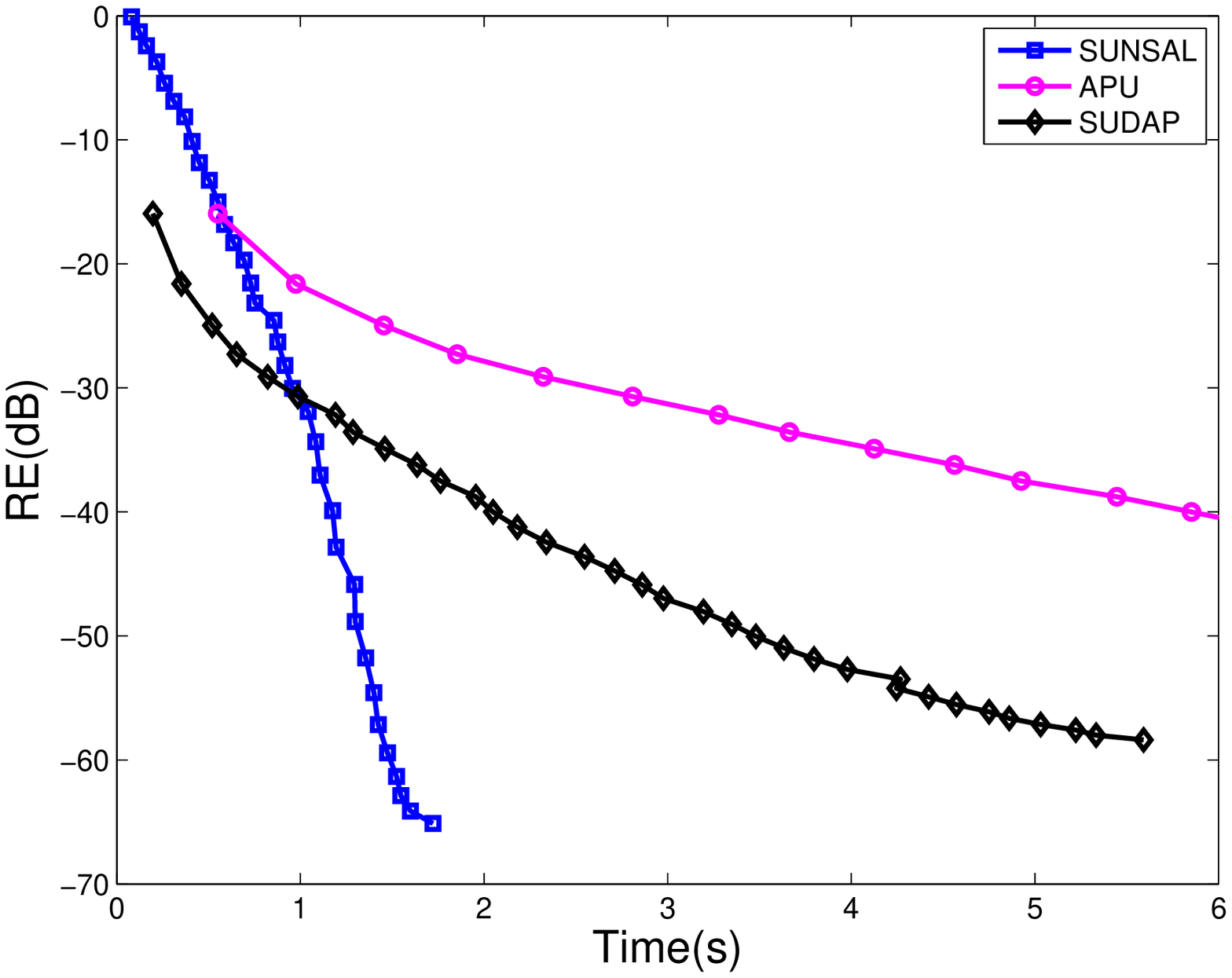}
\caption{Objective function (left) and RE (right) \emph{vs.} time for SUNSAL, APU and SUDAP (Cuprite data).}
\label{fig:Real_Obj_Cuprite}
\end{figure}

From Fig. \ref{fig:HS_Cuprite} (right), the signatures appear to be highly correlated, which makes the unmixing quite challenging. This can be confirmed by computing the smallest angle between any couple of endmembers, which is equal to $\alpha=2.46$ (in degree).
This makes the projection-based methods, including SUDAP and APU, less efficient
since alternating projections are widely known for their slower convergence when the convex
sets exhibit small angles, which is consistent with the convergence analysis in Section \ref{subsec:conv_als}. 
Fig. \ref{fig:Real_Obj_Cuprite}, which depicts the objective function and the RE w.r.t. the computational times 
corroborates this point. Indeed, SUDAP performs faster than SUNSAL and APU if the algorithms are stopped before
RE$<-30$dB. For lower RE$_t$, SUNSAL surpasses SUDAP.

%% file: appendix_A.tex
\noindent Following the KKT conditions, the problem \eqref{eq:opt_ASC_ANC_a}
can be reformulated as finding $\bfu^\ast$ satisfying the following conditions
\begin{equation}
\begin{array}{rcl}
\bfu^\ast- \bfz + \mu \bfb - \lambda\bfd_i  &=& \bs{0}\\
\bfd_i^T \bfu^\ast &\geq &0\\
\bfb^T\bfu^\ast&=&1\\
\lambda& \geq &0\\
\mu& \geq &0\\
\lambda \bfd_i^T \bfu^{\ast} &=& 0.
\end{array}
\end{equation}
Direct computations lead to
\begin{equation}
\bfu^\ast= \bfz-\tilde{\bfz}+ \Delta \bfz
\label{eq:opt_u}
\end{equation}
where
\begin{equation}\label{eq:def_Ustar2}
\begin{array}{rcl}
\tilde{\bfz}& =& \bfc \left(\bfb^T\bfz  -1 \right) \\
\bfc&= & \bfb / \left\|\bfb\right\|^2_2\\
\Delta \bfz &=&\tau_{i} \bfs_i\\
\tau_{i} & =& \max\{0,-\bfd_i^T \left(\bfz-\tilde{\bfz}\right) / \|\bfP \bfd_i\|_2\} \\
\bfs_i &=& \bfP \bfd_i / \|\bfP \bfd_i\|_2\\
\bfP &=&\Id{m}- \bfb\bfb^T / \left\|\bfb\right\|^2_2.\\
\end{array}
\end{equation}

Computing the projection of $\bfz_j$ for $j=1,\cdots,n$  
can be conducted in parallel, leading to the following matrix update rule
\begin{equation}
\begin{array}{ll}
\bfU_i^\ast &= \Pi_{\calS \cap \calN_i}(\bfZ)\\
&=\bfZ-\tilde{\bfZ} + \bfs_i \bs{\tau}_i^T\\
&=\Pi_{\calS}{(\bfZ)} + \bfs_i \bs{\tau}_i^T
\end{array}
\label{eq:proj_kkt}
\end{equation}
where
\begin{equation*}
\begin{split}
&\tilde{\bfZ} =  \bfc \left(\bfb^T \bfZ -\bs{1}_n^T \right) \\
&\bs{\tau}_i^T  = \max\{{\bf 0}, -\bfd_i^T \left(\bfZ-\tilde{\bfZ}\right) / \|\bfP \bfd_i\|_2\}.\\
\end{split}
\end{equation*}
As a conclusion, the updating rules \eqref{eq:proj_kkt} and \eqref{eq:Proj_Geometry} 
only differ by the way the projection $\Pi_{\calS}{(\bfZ)}$ onto $\calS$ has been computed.
However, it is easy to show that $\Pi_{\calS}{(\bfZ)} = \bfZ-\tilde{\bfZ}$ used 
in \eqref{eq:proj_kkt} is fully equivalent to $\Pi_{\calS}{(\bfZ)} = {\bf c}{\bf 1}_n^T + {\bf P}(\bfZ - {\bf c}{\bf 1}_n^T)$
required in \eqref{eq:Proj_Geometry}.


\begin{remark*}
It is worthy to provide an alternative geometric interpretation of the KKT-based solution \eqref{eq:opt_u}.
First, $\bfz-\tilde{\bfz}$ is the projection of $\bfz$
onto the affine set $\calS$. Second, if the projection is inside
the set $\calN_i$, which means $\bfd_i^T \left(\bfz-\tilde{\bfz}\right) \geq 0$,
then the projection of $\bfz$ onto the intersection $\calS \cap \calN_i$
is $\bfz-\tilde{\bfz}$. If the projection is outside of the set $\calN_i$,
implying that $\bfd_i^T \left(\bfz-\tilde{\bfz}\right) < 0$,
a move $\Delta \bfz $ inside the affine set $\calS$ should be added to
$\bfz-\tilde{\bfz}$ to reach the set $\calN_i$.
This move $\Delta \bfz$ should ensure three constraints:
1) $\Delta \bfz$ keeps the point $\bfz-\tilde{\bfz}+\Delta \bfz$ inside the affine set $\calS$,
2) $\bfz-\tilde{\bfz}+\Delta \bfz$ is on the boundary of the set $\calN_i$, and 
3) the Euclidean norm of $\Delta \bfz$ is minimal.
The first constraint, which can be formulated as $\bfb^T\Delta \bfz =0$, is ensured by imposing a move of the form $\Delta \bfz = \bfP \bfw$ where $\bfP=\bfV\bfV^T$ is the projector onto the subspace $\calS_0$ orthogonal to $\bfb$. The second constraint is fulfilled when $\bfd_i^T \left(\bfz-\tilde{\bfz}+\Delta \bfz\right)=0$,
leading to ${\bfd_i^T} {\bfP\bfw} = -\delta_{i}$, where $\delta_{i}= \bfd_i^T \left(\bfz-\tilde{\bfz}\right)$. Thus, due to the third constraint, $\bfw$ can be defined as
\begin{equation}
\bfw = \operatornamewithlimits{argmin}_{\bf v} \|{\bf Pv}\|_2^2\ \textrm{ s.t. } \ \bfd_i^T \bfP \bfw = -\delta_{i}.
\end{equation}
Using the fact that $\bfP$ is an idempotent matrix, i.e., $\bfP^2=\bfP$, the constrained optimization problem
can be solved analytically with the method of Lagrange multipliers, leading to
\begin{equation}
\bfw={-\delta_{i}}{\left(\bfd_i^T \bfP \bfd_i\right)}^{-1} \bfd_i
\end{equation}
and  $\Delta \bfz={\bfP \bfw}={-\delta_{i}}{\left(\bfd_i^T \bfP \bfd_i\right)}^{-1} \bfP \bfd_i$. This final result is consistent with the move defined in \eqref{eq:opt_u} and \eqref{eq:def_Ustar2} by setting $\tau_{i}= \max\{0,-\frac{\delta_{i}}{\|\bfP \bfd_i\|_2}\}$ and $\bfs_i=\bfP \bfd_i / \|\bfP \bfd_i\|_2$. Recall that $\left\|\bfP \bfd_i\right\|_2^2 = \left(\bfd_i^T \bfP \bfd_i\right)$ since $\bfP^T\bfP = \bfP$.
\end{remark*}